# Flexibly Instructable Agents

**Scott B. Huffman**                                         HUFFMAN@TC.PW.COM
*Price Waterhouse Technology Centre, 68 Willow Road*
*Menlo Park, CA 94025 USA*

**John E. Laird**                                            LAIRD@EECS.UMICH.EDU
*Artificial Intelligence Laboratory*
*The University of Michigan, 1101 Beal Ave.*
*Ann Arbor, MI 48109–2110 USA*

## Abstract

This paper presents an approach to learning from situated, interactive tutorial instruction within an ongoing agent. Tutorial instruction is a flexible (and thus powerful) paradigm for teaching tasks because it allows an instructor to communicate whatever types of knowledge an agent might need in whatever situations might arise. To support this flexibility, however, the agent must be able to learn multiple kinds of knowledge from a broad range of instructional interactions. Our approach, called *situated explanation*, achieves such learning through a combination of analytic and inductive techniques. It combines a form of explanation-based learning that is situated for each instruction with a full suite of contextually guided responses to incomplete explanations. The approach is implemented in an agent called INSTRUCTO-SOAR that learns hierarchies of new tasks and other domain knowledge from interactive natural language instructions. INSTRUCTO-SOAR meets three key requirements of flexible instructability that distinguish it from previous systems: (1) it can take known or unknown commands at any instruction point; (2) it can handle instructions that apply to either its current situation or to a hypothetical situation specified in language (as in, for instance, conditional instructions); and (3) it can learn, from instructions, each class of knowledge it uses to perform tasks.

## 1. Introduction

The intelligent, autonomous agents of the future will be called upon to perform a wide and varying range of tasks, under a wide range of circumstances, over the course of their lifetimes. Performing these tasks requires knowledge. If the number of possible tasks and circumstances is large and variable over time (as it will be for a general agent), it becomes nearly impossible to preprogram all of the knowledge required. Thus, knowledge must be added during the agent's lifetime. Unfortunately, such knowledge cannot be added to current intelligent systems while they perform; they must be shut down and programmed for each new task.

This work examines an alternative: intelligent agents that can be *taught* to perform tasks through tutorial instruction, as a part of their ongoing performance. Tutorial instruction is a highly interactive dialogue that focuses on the specific task(s) being performed. While working on tasks, a student may receive instruction as needed to complete tasks or to understand aspects of the domain or of previous instructions. This situated, interactive form of instruction produces very strong human learning (Bloom, 1984). Although it has received little attention in AI, it has the potential to be a powerful knowledge source for artificial agents as well.

Much of tutorial instruction's power comes from its *communicative flexibility*: The instructor can communicate whatever type of knowledge a student may need in whatever situation it is needed. The challenge in designing a tutorable agent is to support the breadth of interaction and learning abilities required by this flexible communication.

In this paper, we present a theory of learning from tutorial instruction within an ongoing agent. In developing the theory, we have given special attention to supporting communicative flexibility for the instructor (the human user). We began by identifying the properties of tutorial instruction from the instructor's perspective. From these properties, we have derived a set of requirements that an instructable agent must meet to support flexible instructability. These requirements drove the development of the theory and its evaluation. Finally, we have implemented the theory in an instructable agent called INSTRUCTO-SOAR (Huffman, 1994; Huffman & Laird, 1993, 1994), and evaluated its performance.[1]

Identifying requirements for flexible instructability provides a target – a set of evaluation criteria – for instructable agents. The requirements encompass the ways an agent interacts with its instructor, comprehends instructions, and learns from them. The most general requirements are common to all interactive learning systems; e.g., the agent is expected to learn general knowledge from instructions, to learn quickly (with a minimal number of examples), to integrate what is learned with its previous knowledge, etc. Other requirements are specific to tutorial instruction.

Our theory of learning from tutorial instruction specifies how analytic and inductive learning techniques can be combined within an agent to meet the requirements, producing general learning from a wide range of instructional interactions. We present a learning framework called *situated explanation* that utilizes the situation an instruction applies to and the larger instructional context (the instruction's type and place in the current dialogue) to guide the learning process. Situated explanation combines a form of explanation-based learning (DeJong & Mooney, 1986; Mitchell, Keller, & Kedar-Cabelli, 1986) that is situated for each individual instruction, with a full suite of contextually guided responses to incomplete explanations. These responses include delaying explanation until more information is available, inducing knowledge to complete explanations, completing explanations through further instruction, or abandoning explanation in favor of weaker learning methods. Previous explanation-based learning systems have employed one or in some cases a static sequence of these options, but have not chosen dynamically among all the options based on the context of each example. Such dynamic selection is required for flexible instructability. The learning framework is cast within a computational model for general intelligent behavior called the *problem space computational model*.

INSTRUCTO-SOAR is an implemented agent that embodies the theory. From interactive natural language instructions, INSTRUCTO-SOAR learns to perform new tasks, extends known tasks to apply in new situations, and acquires a variety of other types of domain knowledge. It allows more flexible instruction than previous instructable systems (e.g., learning apprentice systems, Mitchell, Mahadevan, & Steinberg, 1990) by meeting three key requirements of tutorial instruction: (1) it can take known or unknown commands at any instruction point; (2) it can handle instructions that apply to either its current situation or to a hypothetical situation specified in language (as in, for instance, conditional instructions); and (3) it can learn, from instructions, each class of knowledge it uses to perform tasks.

---

1. Because our work is inspired by human students, we have also taken cues from psychological effects where appropriate. The theory's potential as a cognitive model is discussed elsewhere (Huffman, 1994; Huffman, Miller, & Laird, 1993).

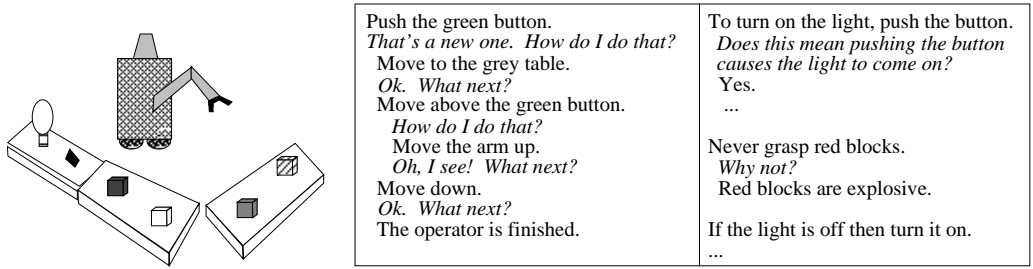

| Push the green button. | To turn on the light, push the button. |
| *That's a new one. How do I do that?* | *Does this mean pushing the button* |
| Move to the grey table. | *causes the light to come on?* |
| *Ok. What next?* | Yes. |
| Move above the green button. | ... |
| *How do I do that?* | |
| Move the arm up. | Never grasp red blocks. |
| *Oh, I see! What next?* | *Why not?* |
| Move down. | Red blocks are explosive. |
| *Ok. What next?* | |
| The operator is finished. | If the light is off then turn it on. |
| | ... |

Figure 1: An example of tutorial instruction.

In what follows, we first discuss the properties and requirements of tutorial instruction. Then, we present our approach and its implementation in INSTRUCTO-SOAR, including a series of examples illustrating the instructional capabilities that are supported. We conclude with a discussion of limitations and areas for further research.

## 2. Properties of Tutorial Instruction

Tutorial instruction is *situated*, *interactive* instruction given to an agent as it attempts to perform tasks. It is *situated* in that it applies to particular task situations that arise in the domain. It is *interactive* in that the agent may request instruction as needed. This type of instruction is common in task-oriented dialogues between experts and apprentices (Grosz, 1977). An example of tutorial instruction given to INSTRUCTO-SOAR in a robotic domain is shown in Figure 1.

Tutorial instruction has a number of properties that make it flexible and easy for the instructor to produce:

P1. **Situation specificity.** Instructions are given for particular tasks in particular situations. To teach a task, the instructor need only provide suggestions for the specific situation at hand, rather than producing a global procedure that includes general conditions for applicability of each step, that handles all possible contingencies, etc. The situation can also help to disambiguate an otherwise ambiguous instruction. A number of authors have discussed the advantages of situation-specific knowledge elicitation (e.g., Davis, 1979; Gruber, 1989).

P2. **Situation specification as needed.** Although instructions typically apply to the situation at hand, the instructor is free to specify other situations as needed; for instance, specifying contingencies using conditional instructions.

P3. **Incremental as-needed elicitation.** Knowledge is elicited incrementally as a part of the agent's ongoing performance. Instructions are given when the agent is unable to perform a task; thus, they directly address points where the agent's knowledge is lacking.

P4. **Task structuring by instructor.** The instructor can structure larger tasks into smaller subtasks in any way desired. For instance, a task requiring ten primitive steps may be taught as a simple sequence of the ten steps, or as two subtasks of five steps each, etc. If the agent does not know what an instructed action or subtask is, or how to perform it in the situation at hand, it will ask for further instruction.

P5. **Knowledge-level interaction.** The instructor provides knowledge to the agent at the knowledge level (Newell, 1981). That is, the instructions refer to objects and actions in the world, not to symbol-level structures (e.g., data structures) within the agent. The interaction occurs in natural language, the language that the instructor uses to talk about the task, rather than requiring artificial terminology and syntax to specify the agent's internal data and processes.

Tutorial instructions provide knowledge applicable to the agent's current situation or a closely related one. Thus, this type of instruction is most appropriate for tasks with a *local control structure*, in which control decisions are made based on presently available information. Local control structure is characteristic of constructive synthesis tasks, in which primitive steps are composed one after another to form a complete solution. Our work focuses on this type of task.[2]

## 3. Requirements on an Instructable Agent

Although easing the instructor's burden in providing knowledge, the properties of tutorial instruction described above place severe requirements on an instructable agent. In general, such an agent must solve three conceptually distinct problems: it must (1) comprehend individual instructions to produce behavior, (2) support a flexible dialogue with its instructor, and (3) produce general learning from the interaction. The properties of tutorial instruction described in the previous section place requirements on the solutions to each of these problems. In what follows, we identify key requirements for each problem in turn.

### 3.1 Comprehending Instructions: The Mapping Problem

The mapping problem involves comprehending instructions that are given in natural language and transforming the information they contain into the agent's internal representation language. This is required for the agent to apply information communicated by instructions at the knowledge level (property P5, above) to its internal processing.

---

2. In contrast, problem solving methods like constraint satisfaction and heuristic classification involve global control strategies. These strategies either follow a fixed global regime or require an aggregation of information from multiple problem solving states to make control decisions. It is possible to produce a global control strategy using a combination of local decisions (Yost & Newell, 1989). However, *teaching* a global method by casting it purely as a sequence of local decisions may be difficult. Other types of instruction, beyond the scope of this work, are required to teach global methods in a natural way. To acquire knowledge for tasks that involve a *known* global control strategy, it may be most efficient to use a method-based knowledge acquisition tool (e.g., Birmingham & Klinker, 1993; Birmingham & Siewiorek, 1989; Eshelman, Ehret, McDermott, & Tan, 1987; Marcus & McDermott, 1989; Musen, 1989) with that control strategy built in.

Solving the mapping problem in general involves all of the complexities of natural language comprehension. As Just and Carpenter (1976) point out, instructions can be linguistically complex and difficult to interpret independent of the difficulty of the task being instructed. Even in linguistically simple instructions, actions and objects are often incompletely specified, requiring the use of context and domain knowledge to produce a complete interpretation (Chapman, 1990; DiEugenio & Webber, 1992; Frederking, 1988; Martin & Firby, 1991).

The general requirement for the mapping problem on a tutorable agent is straightforward:

$M_1$. A tutorable agent must be able to comprehend and map all aspects of each instruction that fall within the scope of information it can possibly represent.

The agent cannot be expected to interpret aspects that fall outside its representation abilities (these abilities may be augmented through instruction, but this occurs by building up from existing abilities). A more detailed analysis could break this general requirement into a set of more specific ones.

This work has not focused on the mapping problem. Rather, the agent we have implemented uses fairly standard natural language processing techniques to handle instructions that express a sufficient range of actions and situations to demonstrate its other capabilities. We have concentrated our efforts on the interaction and transfer problems.

## 3.2 Supporting Interactive Dialogue: The Interaction Problem

The interaction problem is the problem of supporting flexible dialogue with an instructor. The properties of tutorial instruction indicate that this dialogue occurs during the agent's ongoing performance to address its lacks of knowledge (property P3); within the dialogue, the agent must handle instructions that apply to different kinds of situations (properties P1 and P2) and that structure tasks in different ways (property P4).

An instructable agent moves toward solving the interaction problem to the degree that it supports these properties. In this work, we concentrate on the *instructor's* utterances within the dialogue, since flexibility for the instructor is the goal. We have not considered the potential complexity of the agent's utterances (e.g., to give the instructor various kinds of feedback) in much detail.

The properties of flexible interaction can be specified in terms of individual instruction events, where an instruction event is the utterance of a single instruction at a particular point in the discourse. To support truly flexible dialogue, an instructable agent must be able to handle any instruction event that is coherent at the current discourse point. Each instruction event is initiated by either the student or the teacher, and carries knowledge of some type to be applied to a particular task situation. Thus, a flexible tutorable agent should support instruction events with:

$I_1$. **Flexible initiation.** Instruction events can be initiated by agent or instructor.

$I_2$. **Flexibility of knowledge content.** The knowledge carried by an instruction event can be any piece of any of the types of knowledge the agent uses that is applicable in some way within the ongoing task and discourse context.

$I_3$. **Situation flexibility.** An instruction event can apply either to the current task situation or to some specified hypothetical situation.

The following sections discuss each of these requirements in more detail.

### 3.2.1 FLEXIBLE INITIATION

In human tutorial dialogues, initiation of instruction is mixed between student and teacher. One study indicates that teacher initiation is more prevalent early in instruction; student initiation increases as the student learns more, and then drops off again as the student masters the task (Emihovich & Miller, 1988).

*Instructor*-initiated instruction is difficult to support because instruction events can interrupt the agent's ongoing processing. Upon interrupting the agent, an instruction event may alter the agent's knowledge in a way that could change or invalidate the reasoning in which the agent was previously engaged. Because of these difficulties, instructable systems to date have not fully supported instructor-initiated instruction.[3] Likewise, INSTRUCTO-SOAR does not handle instructor-initiated instruction.

*Agent*-initiated instruction can be directed in (at least) two possible ways: by verification or by impasses. Some learning apprentice systems, such as LEAP (Mitchell et al., 1990) and DISCIPLE (Kodratoff & Tecuci, 1987b) ask the instructor to verify or alter each reasoning step. The advantage of this approach is that each step is examined by the instructor; the disadvantage, of course, is that each step *must* be examined. An alternative approach is to drive instruction requests by impasses in the agent's task performance (Golding, Rosenbloom, & Laird, 1987; Laird, Hucka, Yager, & Tuck, 1990). This is the approach used by INSTRUCTO-SOAR. An impasse indicates that the agent's knowledge is lacking and it needs instruction. The advantage of this approach is that as the agent learns, it becomes more autonomous; its need for instruction decreases over time. The disadvantage is that not all lacks of knowledge can be recognized by reaching impasses; e.g., no impasse will occur when performance is correct but inefficient.

### 3.2.2 FLEXIBILITY OF KNOWLEDGE CONTENT

A flexible tutorable agent must handle instruction events involving any knowledge that is applicable in some way within the ongoing task and discourse context. This requirement is difficult to meet in general, because of the wide range of knowledge that may be relevant to any particular situation. It requires a robust ability to relate each utterance to the ongoing discourse and task situation. No instructable systems have met this requirement fully.

However, we can define a more constrained form of this requirement, limited to instructions that command actions (i.e., imperatives). Imperative commands are especially prevalent in tutorial instruction of procedures. Supporting flexible knowledge content for commands means allowing the instructor to give any relevant command at any point in the dialogue for teaching a task. We call this ability *command flexibility*.

---

3. Some systems have learned purely by observing an expert (e.g., Dent, Boticario, McDermott, Mitchell & Zabowski, 1992; Redmond, 1992). Observation is a type of instructor-initiatedness, but the instruction is not an interactive dialogue.

For any command that is given, there are three possibilities: (1) the commanded action is known, and the agent performs it; (2) the commanded action is known, but the agent does not know how to perform it in the current situation (extra, unknown steps are needed); or (3) the commanded action is unknown. Thus, command flexibility allows the instructor teaching a procedure to skip steps (2) or to command a subtask that is unknown (3) at any point. In such cases, the agent asks for further instruction. The interaction pattern that results, in which procedures are commanded and then taught as needed, has been observed in human instruction. Wertsch (1979) notes that "...adults spontaneously follow a communication strategy in which they use directives that children do not understand and then guide the children through the behaviors necessary to carry out these directives."

Command flexibility gives the instructor great flexibility in teaching a set of tasks because the instructions can hierarchically structure the tasks in whatever way the instructor wishes. A mathematical analysis (Huffman, 1994) revealed that the number of possible sequences of instructions that can be used to teach a given procedure grows exponentially with the number of actions in the procedure. For a procedure with 6 primitive actions, there are over 100 possible instruction sequences; for 7, there are over 400.

### 3.2.3 SITUATION FLEXIBILITY

A flexible tutorable agent must handle instructions that apply to either the current task situation or some hypothetical situation that the instructor specifies. Instructors make frequent use of both of these options. For instance, analysis of a protocol of a student being taught to use a flight simulator revealed that 119 out of 508 instructions (23%) involved hypothetical situations, with the remainder applying to the current situation at the time they were given.

Instructions that apply to the current situation, such as imperative commands (e.g., "Move to the yellow table"), are called *implicitly situated* (Huffman & Laird, 1992). Since the instruction itself says nothing about the situation to which it should be applied, the current situation (the task being performed and the current state) is implied.

In contrast, instructions that specify elements of the situation to which they are meant to apply are *explicitly situated* (Huffman & Laird, 1992). The agent is not meant to carry out these instructions immediately (as an implicitly situated instruction), but rather when a situation arises that is like the one specified. Examples include conditionals and instructions with purpose clauses (DiEugenio, 1993), such as the following:[4]

- *When using chocolate chips*, add them to coconut mixture just before pressing into pie pan.

- *To restart this*, you can hit R or shift-R.

- *When you get to the interval that you want*, you just center up the joystick again.

---

4. These examples are taken from a protocol of tutorial instruction and a written source of instruction (a cookbook).

As a number of researchers have pointed out (Ford & Thompson, 1986; Haiman, 1978; Johnson-Laird, 1986), conditional clauses introduce a shared reference between speaker and hearer that forms an explicit background for interpreting or evaluating the consequent.[5] Here, the clauses in italics indicate a hypothetical situation to which the command in the remainder of the instruction is meant to apply. In most cases, the situation is only partially specified, with the remainder drawn from the current situation, as in "When using chocolate chips [and cooking this recipe, and at the current point in the process]..."

In general, a hypothetical situation may be created and referred to across multiple utterances. The agent presented here handles both implicitly and single explicitly situated instructions, but does not deal with hypothetical situations that exist through multiple instructions.

### 3.3 Producing General Learning: The Transfer Problem

The transfer problem is the problem of learning generally applicable knowledge from instructions, that will transfer to appropriate situations in the future. This general learning is based on instructions that apply to specific situations (property P1, above). Many types of knowledge may be learned, since instructions can provide any type of knowledge that the agent is lacking (property P3).

Solving this problem involves more than simply memorizing instructions for future use; rather, conditions for applying each instruction must be determined from the situation. Consider, for example, the following exchange between instructor and agent:

> Block open our office door.
> *How do I do that?*
> Pick up a red block.
> Now, drop it here, next to the door.

What are the proper conditions for performing the "pick up" action? Simple memorization yields poor learning; e.g., `whenever blocking open an office door, pick up a red block`. However, the block's color, and even the fact that it is a block, are irrelevant in this case. Rather, the fact that the block weighs (say) more than five pounds, giving it enough friction with the floor to hold open the door, is crucial. Thus, the proper learning might be:

> **If**   trying to block open a door, and
>    there is an object **obj** that is can be picked up, and
>    **obj** weighs more than 5 pounds
> **then** propose picking up **obj**.

Here, the original instruction is both generalized (`color red` and `isa block` drop out) and specialized (`weight > 5` is added).

The transfer problem places a number of demands on a tutorable agent:

$T_1$. **General learning from specific cases.** The agent is instructed in a particular situation, but is expected to learn general knowledge that will apply in sufficiently similar situations.

---

5. Some types of conditionals do not follow this pattern (Akatsuka, 1986), but they are not relevant to tutorial instruction.

$T_2$. **Fast learning.** An instructable agent is expected to learn new procedures quickly. Typically, a task should only have to be taught once.

$T_3$. **Maximal use of prior knowledge.** An agent must apply its prior knowledge in learning from instruction. This is a maxim for machine learning systems in general (if you have knowledge, use it), and is particularly relevant for learning from instruction because learning is expected to happen quickly.

$T_4$. **Incremental learning.** The agent must be able to continually increase in knowledge through instruction. New knowledge must be smoothly integrated with the agent's existing knowledge as a part of its ongoing performance.

$T_5$. **Knowledge-type flexibility.** Since any type of knowledge (e.g., control knowledge, causal knowledge, etc.) might be communicated by instructions, a flexible tutorable agent must be able to learn each type of knowledge it uses. We make this a testable criterion below by laying out the types of knowledge in an agent based on a particular computational model.

$T_6$. **Dealing with incorrect knowledge.** The agent's knowledge is clearly incomplete (otherwise, it would not need instruction); it may also be incorrect. A general tutorable agent must be able to perform and learn effectively despite incorrect knowledge.

$T_7$. **Learning from instruction coexisting with learning from other sources.** In addition to instruction, a complete agent should be able to learn from other sources of knowledge that are available. These might include learning from observation/demonstrations, experimentation in the environment, analogy, etc.

The theory of learning from tutorial instruction presented here focuses on extending incomplete knowledge through instruction – requirements $T_1$ through $T_5$ of this list. Handling incorrect knowledge ($T_6$) and learning from other sources ($T_7$) are planned extensions in progress.

Table 1 summarizes the requirements that must be met by an instructable agent to support flexible tutorial instruction, and indicates the requirements targeted by Instructo-Soar. We have made two simplifications in using the requirements to evaluate Instructo-Soar. First, we treat each requirement as *binary*; that is, as if either completely met or unmet. In reality, some requirements could be broken into finer-grained pieces to be evaluated separately. Second, we treat each requirement *independently*. The table indicates Instructo-Soar's performance on each requirement alone, but does not account for potential interactions between them. These interactions can be complex; for instance, in pursuing fast learning ($T_2$), an agent may sacrifice good general learning ($T_1$) because it bases its generalizations on too few examples. We have not addressed such tradeoffs in our evaluation of Instructo-Soar.

# 4. Related Work

Although there has not been extensive research on agents that learn from tutorial instruction *per se*, learning from instruction-like input has been a long-time goal in AI (Carbonell,

| Problem | | Requirement | Instructo-Soar? | |
|---|---|---|---|---|
| Mapping | $M_1$ | Mapping of all representable information | no | (as needed to show other capabilities) |
| Interaction | $I_1$ | Flexible initiation of instruction | no | (only agent-initiated) |
| | $I_2$ | Flexibility of instructional knowledge content | partial | (command flexibility) |
| | $I_3$ | Situation flexibility<br>• implicitly situated<br>• explicitly situated single utterance<br>• explicitly situated multiple utterance | partial<br>yes<br>yes<br>no | |
| Transfer | $T_1$ | General learning from specific cases | yes | (via situated explanation) |
| | $T_2$ | Fast learning | yes | (new procedures only taught once) |
| | $T_3$ | Maximal use of prior knowledge | yes | |
| | $T_4$ | Incremental learning | yes | |
| | $T_5$ | Knowledge-type flexibility | yes | (learns all PSCM knowledge types) |
| | $T_6$ | Ability to deal with incorrect knowledge | no | (only extending incomplete knowledge) |
| | $T_7$ | Learning from instruction coexisting with learning from other sources | no | (not demonstrated) |

Table 1: The requirements on a flexible tutorable agent, and Instructo-Soar's performance on them.

Michalski, & Mitchell, 1983; McCarthy, 1968; Rychener, 1983). Early non-interactive systems learned declarative, ontological knowledge from language (Haas & Hendrix, 1983; Lindsay, 1963), simple tasks from unsituated descriptions (Lewis, Newell, & Polk, 1989; Simon, 1977; Simon & Hayes, 1976), and task heuristics from non-operational advice (Hayes-Roth, Klahr, & Mostow, 1981; Mostow, 1983).

Other work has concentrated on behaving based on interactive natural language instructions. SHRDLU (Winograd, 1972) performed natural language commands and did a small amount of rote learning − e.g., learning new goal specifications by directly transforming sentences into state descriptions. More recent systems that act in response to language (concentrating on the mapping problem) but do only minimal learning include SONJA (Chapman, 1990), AnimNL (DiEugenio & Webber, 1992), and Homer (Vere & Bickmore, 1990).

Some recent work has focused more on learning from situated natural language instructions. Martin and Firby (1991) discuss an approach to interpreting and learning from elliptical instructions (e.g., "Use the shovel") by matching the instruction to expectations generated from a task execution failure. Alterman *et al.*'s FLOBN (Alterman, Zito-Wolf, & Carpenter, 1991; Carpenter & Alterman, 1994) searches for instructions in its environment in order to operate devices. FLOBN learns by relating a device's instructions to known procedures for operating similar devices. These systems do not target learning from flexible interactive instructions or types of instructions other than imperatives, however.

The bulk of work on learning from instruction-like input has been under the rubric of learning apprentice systems (LASs), and closely related programming-by-demonstration (PbD) systems (Cypher, 1993) – as employed, for instance, in recent work on learning within software agents (Dent et al., 1992; Maes, 1994; Maes & Kozierok, 1993; Mitchell, Caruana, Freitag, McDermott, & Zabowski, 1994). These systems learn by interacting with an expert; either observing the expert solving problems (Cypher, 1993; Donoho & Wilkins, 1994; Mitchell et al., 1990; Redmond, 1992; Segre, 1987; VanLehn, 1987; Wilkins, 1990), or attempting to solve problems and allowing the expert to guide and critique decisions that are made (Golding et al., 1987; Gruber, 1989; Kodratoff & Tecuci, 1987b; Laird et al., 1990; Porter, Bareiss, & Holte, 1990; Porter & Kibler, 1986). Each LAS has learned particular types of knowledge: e.g., operator implementations (Mitchell et al., 1990), goal decomposition rules (Kodratoff & Tecuci, 1987b), operational versions of functional goals (Segre, 1987), control knowledge and control features (Gruber, 1989), procedure schemas (a combination of goal decomposition and control knowledge) (VanLehn, 1987), useful macro-operations (Cypher, 1993), heuristic classification knowledge (Porter et al., 1990; Wilkins, 1990), etc.

Tutorial instruction is a more flexible type of instruction than that supported by past LASs, for three reasons. First, the instructor may command unknown tasks or tasks with unachieved preconditions to the agent at any instruction point (command flexibility). Past LASs limit input to particular commands/observations at particular times (e.g., only commanding or observing directly executable actions) and typically do not allow unknown commands at all. Second, tutorial instruction allows the use of explicitly situated instructions (situation flexibility), to teach about contingencies that may not be present in the current situation; past LASs do not. Third, tutorial instruction requires that all types of task knowledge can be learned (knowledge-type flexibility). Past LASs learn only a subset of the types of knowledge they use to perform tasks.

## 5. A Theory of Learning from Tutorial Instruction

Our theory of learning from tutorial instruction consists of a learning framework, *situated explanation*, placed within a computational model for general agenthood, the *problem space computational model*. We first describe the computational model and then the learning framework.

### 5.1 The Problem Space Computational Model

A computational model (CM) is a "a set of operations on entities that can be interpreted in computational terms" (Newell *et al.*, 1990, p. 6). A computational model for a general instructable agent must meet two requirements:

1. **Support of general computation/agenthood.**

2. **Close correspondence to the knowledge level.** Because tutorial instructions provide knowledge at the knowledge level (Newell, 1981), the further the CM com-ponents are from the knowledge level, the more difficult mapping and learning from instructions will be. In addition, a close correspondence to the knowledge level will allow us to use the CM to identify the types of knowledge the agent uses.

Many potential CMs are ruled out by these requirements. Standard programming languages (e.g., Lisp) and theoretical CMs like Turing machines and push-down automata support general computation, but their operations and constructs are at the symbol level, without direct correspondence to the knowledge level. Similarly, connectionist and neural network models of computation (e.g., Rumelhart & McClelland, 1986) employ (by design) computational operations and entities at a level far below the knowledge level. Thus, these models are not appropriate as the top-level CM for an instructable agent. However, because higher levels of description of a computational system are implemented by lower levels (Newell, 1990), these CMs might be used as the implementation substrate for the higher level CM of an instructable agent.

Another alternative is logic, which has entities that are well matched to the knowledge level (e.g., propositions, well-formed formulas). Logics specify the set of legal operations (e.g., *modus ponens*), thus defining the space of what can *possibly* be computed. However, logic provides no notion of what *should* be computed; that is, logics alone do not specify the *control* of the logical operations' application. It is desirable that the CM of an instructable agent include control knowledge, because control knowledge is a crucial type of knowledge for general agenthood, that can be communicated by instructions.

Since one of our goals is to identify an agent's knowledge types, it might appear that selecting a theory of knowledge representation would be more appropriate than selecting a computational model. Such theories define the functions and structures used to represent knowledge (e.g., KL-ONE, Brachman, 1980); some also define the possible content of those structures (e.g., conceptual dependency theory, Schank, 1975; CYC, Guha & Lenat, 1990). However, computational structure must be added to these theories to produce working agents. Thus, rather than an *alternative* to specifying a computational model, a theory of knowledge representation is an *addition*. A content theory of knowledge representation would provide a more fine-grained breakdown of the knowledge to be learned by an instructable agent within each category of knowledge specified by its CM. We have not employed a particular content theory in this work thus far, however.

The computational model adopted here is called the problem space computational model (PSCM) (Newell et al., 1990; Yost, 1993). The PSCM is a general formulation of computation in a knowledge-level agent, and many applications have been built within it (Rosenbloom, Laird, & Newell, 1993a). It specifies an agent in terms of computation within problem spaces, without reference to the symbol level structures used for implementation. Because its components approximate the knowledge level (Newell et al., 1990), the PSCM is an apt choice for identifying an agent's knowledge types. Soar (Laird, Newell, & Rosenbloom, 1987) is a symbol level implementation of the PSCM.

A schematic of a PSCM agent is shown in Figure 2. Perception and motor modules connect the agent to the external environment. A PSCM agent reaches a goal by moving through a sequence of *states* in a *problem space*. It progresses toward its goals by sequentially applying *operators* to the current state. Operators transform the state, and may produce motor commands. In PSCM, operators can be more powerful than simple STRIPS operators (Fikes, Hart, & Nilsson, 1972), because they can perform arbitrary computation (e.g., they can include conditional effects, multiple substeps, reactivity to different situations, etc.). The PSCM agent reaches an *impasse* when its immediately available knowledge is not sufficient either to select or fully apply an operator. When this occurs, another problem space context − a *subgoal* − is created, with the goal of resolving the impasse. This second context may impasse as well, causing a third context to arise, and so on.

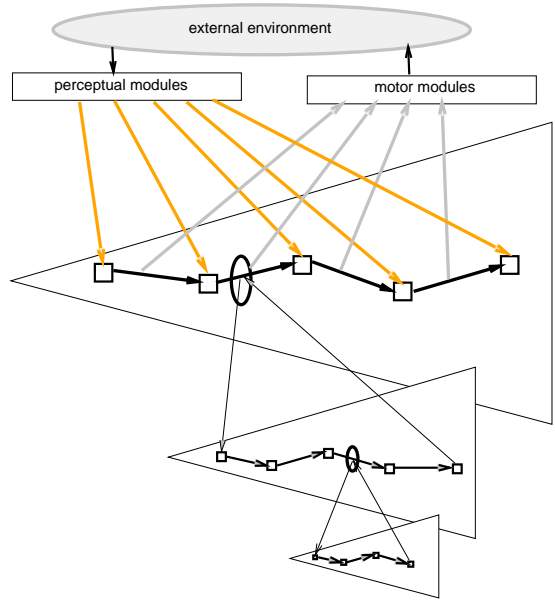

Figure 2: The processing of a PSCM-based agent. Triangles represent problem spaces; squares, states; arrows, operators; and ovals, impasses.

The only computational entities in the PSCM mediated by the agent's knowledge are states and operators. There are a small set of basic PSCM-level operations on these entities that the agent performs:

1. **State inference.** Simple monotonic inferences that are always to be applied can be made without using a PSCM operator. Such inferences augment the agent's representation of the state it is in by inferring state properties based on other state properties (including those delivered by perception). For instance, an agent might know that a block is held if its gripper is closed and positioned directly above the block.

2. **Operator selection.** The agent must select an operator to apply, given the current state. This process involves two types of knowledge:

   2.1. *Proposal knowledge:* Indicates operators deemed appropriate for the current situation. This knowledge is similar to "precondition" knowledge in simple STRIPS operators.

   2.2. *Control knowledge:* Orders proposed operators; e.g., "A is better than B"; "C is best"; "D is rejected."

3. **Operator application.** Once selected, an operator may be applied directly, or indirectly via a subgoal:

   3.1. *Operator effects.* The operator is applied directly in the current problem space. The agent has knowledge of the effects of the operator on the state and motor commands produced (if any).

   3.2. *Sub-operator selection.* The operator is applied by reaching an impasse and selecting operators in a subgoal. Here, knowledge to apply the operator is selection knowledge (2, above) for the sub-operators.

4. **Operator termination.** An operator must be terminated when its application has been completed. The termination conditions, or goal concept (Mitchell et al., 1986), of an operator indicate the state conditions that the operator is meant to achieve. For example, the termination conditions of `pick-up(blk)` might be that `blk` is held and the arm is raised.[6]

Each of these functions is performed by the agent using knowledge; thus, they define the set of knowledge types present within a PSCM agent. The knowledge types (five types total) are summarized in Table 2. Because Soar is an implementation of the PSCM, all knowledge within Soar agents is of these types.

In Soar's implementation of the PSCM, learning occurs whenever results are returned from a subgoal to resolve impasses. The learning process, called *chunking*, creates rules (called chunks) that summarize the processing in the subgoal leading to the creation of the result. Depending on the type of result, chunks may correspond to any of the five types of PSCM knowledge. When similar situations arise in the future, chunks allow the impasse that caused the original subgoal to be avoided by producing their results directly. Chunking is a form of explanation-based learning (Rosenbloom & Laird, 1986). Although it is a summarization mechanism, through taking both inductive and deductive steps in subgoals, chunking can produce both inductive and deductive learning (Miller, 1993; Rosenbloom & Aasman, 1990). Chunking occurs continuously, making learning a part of the ongoing activity of a Soar/PSCM agent.

The PSCM clarifies the task of an instructable agent: it must be able to learn each of the five types of PSCM knowledge from instruction. The next section discusses the learning process itself.

## 5.2 Learning from Instructions through Situated Explanation

Learning from instruction involves both analytic learning (learning based on prior knowledge) and inductive learning (going beyond prior knowledge). Analytic learning is needed because the agent must learn from instructions that combine known elements – e.g., learning to pick up objects by combining known steps to pick up a particular object. The agent's prior knowledge of these elements can be used to produce better and faster learning. Inductive learning is needed because the agent must learn new task goals and domain knowledge that are beyond the scope of its prior knowledge. The goal of this research is not to produce more powerful analytic or inductive techniques, but rather to specify how these techniques come together to produce a variety of learning in the variety of instructional situations faced by an instructable agent. The resulting approach is called *situated explanation*.

---

6. PSCM operators have explicit termination knowledge because they can have a string of conditional effects that take place over time, they can respond to (or wait for) the external environment, etc. STRIPS operators, in contrast, do not need explicit termination knowledge, because they are defined by a single list of effects, and are "terminated" by definition after applying those effects.

| Entity | Knowledge type | Example |
|---|---|---|
| state | inference | If gripper is closed & directly above obj → holding obj. |
| operator | proposal | If goal is to pick up obj on table-x, and not docked at table-x, then propose moving to table-x. |
| operator | control | If goal is to pick up small metal obj on table-x, prefer moving to table-x over fetching magnet. |
| operator | effects | An effect of the operator move to table-x is that the robot becomes docked at table-x. |
| operator | termination | Termination conditions of pick up obj are that the gripper is raised & holding obj. |

Table 2: The five types of knowledge of PSCM agents.

Instruction requirements $T_1$ through $T_3$ specify that general learning ($T_1$) must occur from single, specific examples ($T_2$), by making maximal use of prior knowledge ($T_3$). These requirements bode strongly for a learning approach based on *explanation*. The use of explanation to produce general learning has been a common theme in machine learning (e.g., DeJong & Mooney, 1986; Fikes et al., 1972; Minton, Carbonell, Knoblock, Kuokka, Etzioni, & Gil, 1989; Rosenbloom, Laird, & Newell, 1988; Schank & Leake, 1989; many others) and cognitive science (Anderson, 1983; Chi, Bassok, Lewis, Reimann, & Glaser, 1989; Lewis, 1988; Rosenbloom & Newell, 1986). Forming explanations enables general learning from specific cases (requirement $T_1$) because the explanation indicates which features of a case are important and which can be generalized. Learning by explaining typically requires only a single example (requirement $T_2$) because the prior knowledge employed to construct the explanation (requirement $T_3$) provides a strong bias that allows this fast learning.

Thus, we use an explanation-based method as the core of our learning from instruction approach, and fall back on inductive methods when explanation fails. In standard explanation-based learning, explaining a reasoning step involves forming a "proof" (using prior knowledge) that the step leads from the current state of reasoning toward the current goal. The proof is a path of reasoning from the current state to the goal, through the step being explained, as diagrammed in Figure 3. General learning is produced by forming a rule that includes only the causally required features of the state, goal, and step appearing in the proof; features that do not appear are generalized away.

Figure 3 indicates the three key elements of an explanation: the step being explained, the endpoints of the explanation (a state $S$ and goal $G$ to be reached), and the other steps required to complete the explanation. What form do these elements of an explanation take for situated explanation of an instruction?

- *Step to be explained.* In situated explanation, the step to be explained is an individual instruction given to the agent.

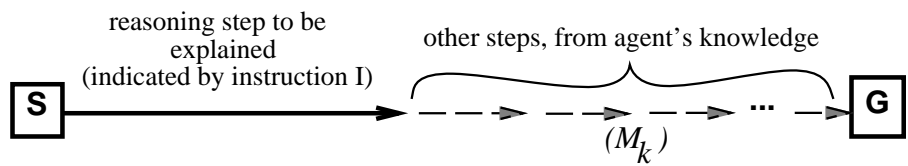

Figure 3: Caricature of an explanation of how a reasoning step applies to a situation starting in a state $S$, with a goal $G$ to be achieved.

Alternatively, an entire instruction episode − e.g., the full sequence of instructions for a new procedure − could be explained at once. Applying explanation to single steps results in knowledge applicable at each step (as in Golding et al., 1987; Laird et al., 1990); explaining full sequences of reasoning steps results in learning schemas that encode the whole reasoning episode (as in Mooney, 1990; Schank & Leake, 1989; Van-Lehn, 1987). Learning factored pieces of knowledge rather than monolithic schemas allows more reactive behavior, since knowledge is accessed locally based on the current situation (Drummond, 1989; Laird & Rosenbloom, 1990). This meshes with the PSCM's local control structure. Explaining individual instructions is also supported by psychological results on the self-explanation effect, which have shown that subjects who self-explain instructional examples do so by re-deriving individual lines of the example. "Students virtually never reflect on the overall solution and try to recognize a plan that spans all the lines" (VanLehn and Jones, 1991, p. 111).

- *Endpoints of explanation.* The endpoints of the explanation − a state $S$ and a goal $G$ to be achieved − correspond to the *situation* that the instruction applies to. Situation flexibility (requirement $I_3$) stipulates that this situation may be either the current state of the world and goal being pursued or some hypothetical situation that is specified explicitly in the instruction. An instruction that does not specify any situational features is implicitly situated, and applies to the agent's current situation. Alternatively, an instruction can specify features of $S$ or $G$, making for two kinds of explicitly situated instructions. For example, "If the light is on, push the button" indicates a hypothetical state with a light on; "To turn on the machine, flip the switch" indicates a hypothetical goal of turning on the machine. A situation $[S, G]$ is produced for each instruction, based on the current task situation and any situation features the instruction specifies.

- *Other required steps.* To complete an explanation of an instruction, an agent must bring its prior knowledge to bear to complete the path through the instruction to achievement of the situation goal. A PSCM agent's knowledge applies to its current situation to select and apply operators and to make inferences. When explaining an instruction $I$, this knowledge is applied internally to the situation $[S, G]$ associated with $I$. That is, explanation takes the form of *forward internal projection* within that situation. As depicted in Figure 3, the agent "imagines" itself in state $S$, and then runs forward, applying the instructed step and any knowledge that it has about subsequent states/operators. This knowledge includes both what is normally used in the external world and knowledge of operators' expected effects that is used to produce those effects in the projected world. If $G$ is reached within the projection, then the projected path from $S$, through the step instructed by $I$, to $G$ comprises an explanation of $I$. By indicating the features of $I$, $S$, and $G$ causally required for success, the explanation allows the agent to learn general knowledge from $I$ (as in standard EBL, realized in our agent by Soar's chunking mechanism, Rosenbloom & Laird, 1986). However, the agent's prior knowledge may be insufficient, causing an incomplete explanation, as described further below.

Combining these elements produces an approach to learning from tutorial instruction that is conceptually quite simple. For each instruction $I$ that is received, the agent first determines what situation $I$ is meant to apply to, and then attempts to explain why the step indicated by $I$ leads to goal achievement in that situation (or prohibits it, for negative instructions). If an explanation can be made, it produces general learning of some knowledge $I_K$ by indicating the key features of the situation and instruction that cause success.

If an explanation cannot be completed, it indicates that the agent is missing one or more pieces of prior knowledge $M_K$ (of any PSCM type) needed to explain the instruction. Missing knowledge (in Figure 3, missing arrows) causes an incomplete explanation by precluding achievement of $G$ in the projection. For instance, the agent may not know a key effect of an operator, or a crucial state inference, needed to reach $G$. More radically, the action commanded by $I$ may be completely unknown and thus inexplicable.

As shown in Figure 4, there are four general options a learning agent might follow when it cannot complete an explanation. ($O1$) It could delay the explanation until later, in the hope that the missing knowledge ($M_K$) will be learned in the meantime. Alternatively, ($O2$-$O3$) it could try to complete the explanation now by somehow learning the missing knowledge. The missing knowledge could be learned ($O2$) inductively (e.g., by inducing over the "gap" in the explanation, as described by VanLehn, Jones & Chi, 1992, and many others), or, ($O3$) in an instructable agent's case, through further instruction. Finally, ($O4$) it could abandon the explanation altogether and try to learn the desired knowledge in another way instead.

Given only an incomplete explanation, it would be difficult to choose which option to follow. Identifying the missing knowledge $M_K$ in the general case is a difficult credit assignment problem (with no algorithmic solution), and there is nothing in the incomplete explanation itself that predicts whether $M_K$ will be learned later if the explanation is delayed. Thus, past machine learning systems have responded to incomplete explanations either in only a single way, or in multiple ways, but that are tried in a fixed sequence. Many authors (Bergadano & Giordana, 1988; Hall, 1988; VanLehn, 1987; VanLehn, Jones, & Chi, 1992; Widmer, 1989), for instance, describe systems that make inductions to complete incomplete explanations (option $O2$). Because of the difficulty of determining missing knowledge, these systems either base their induction on multiple examples, and/or bias the induction with an underlying theory or a teacher's help. SIERRA (VanLehn, 1987), for example, induces over multiple partially explained examples, and constrains the induction by requiring that each of the examples is unexplainable because of the same piece of missing knowledge (the same disjunct, in SIERRA's terminology). SWALE (Schank & Leake, 1989) uses an underlying theory of "anomalies" in explanations to complete incomplete explanations of events. OCCAM (Pazzani, 1991b) uses options $O2$ and $O4$ in a static order:

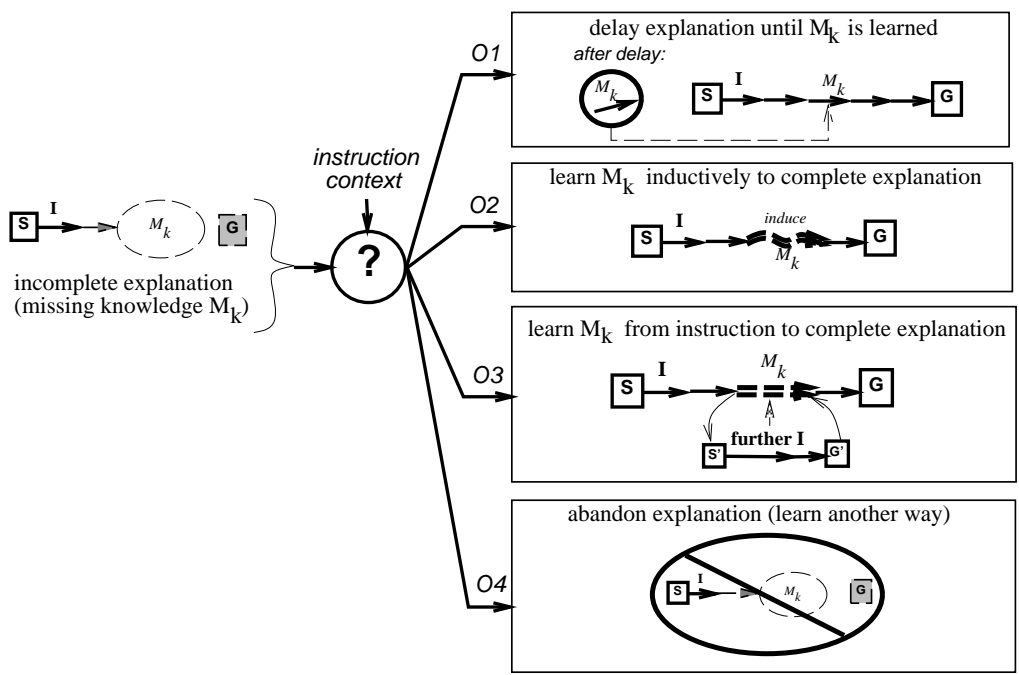

Figure 4: Options when faced with an incomplete explanation because of missing knowledge $M_K$.

It first attempts to fill in the gaps in an incomplete explanation inductively, biased by a naive theory; if that fails, it abandons explanation and falls back on correlational learning methods. PET (Porter & Kibler, 1986) is an example of a system that delays explanation of a reasoning step until it learns further knowledge (option $O1$).

However, as indicated in Figure 4, an instructable agent has additional information available to it besides the incomplete explanation itself. Namely, the *instructional context* (that is, the type of instruction and its place within the dialogue) often indicates which option is most appropriate for a given incomplete explanation. Thus, situated explanation includes all four of the options and dynamically selects between them based on the instructional context. For a situated explanation of an instruction $I$ in a situation $[S, G]$, where missing knowledge $M_K$ precludes completing the explanation to learn knowledge $I_K$, options $O1$-$O4$ take the following form:

$O1$. **Delay the explanation until later.** The instructional context can indicate a likelihood that the missing knowledge $M_K$ will be learned later. For instance, an instruction $I$ given in teaching a new procedure cannot be immediately explained because the remaining steps of the procedure are unknown, but they will be known later (assuming the instructor completes teaching the procedure). In such cases, the agent discards its current, incomplete explanation and simply memorizes $I$'s use in $[S, G]$ (rote learning). Later, after $M_K$ is learned, $I$ is recalled and explained in $[S, G]$, causing $I_K$ to be learned.

---

Given instruction $I$ from which knowledge $I_K$ can be learned:
- Determine the situation $[S, G]$ (current or hypothetical) to which $I$ applies
- Explain $I$ in $[S, G]$ by forward projecting from $S \xrightarrow{I} \rightsquigarrow \cdots \rightsquigarrow G$
  - $\rightarrow$ Success ($G$ met): *learn $I_K$* from the complete explanation ($\approx$ EBL).
  - $\rightarrow$ Failure: missing knowledge $M_K$. Options:
    - $O1$. Delay explanation until later.
    - $O2$. Induce $M_K$, completing the explanation.
    - $O3$. Take instruction to learn $M_K$, completing the explanation.
    - $O4$. Abandon explanation; instead, learn $I_K$ inductively.

---

Table 3: Situated explanation.

$O2$. **Induce $M_K$, completing the explanation.** In some cases, the instructional context localizes the missing knowledge $M_K$ to be part of a particular operator. For instance, a purpose clause instruction ("To do X, do Y") suggests that the single operator Y should cause X to occur. Because this localization tightly constrains the "gap" in the incomplete explanation, the agent can use heuristics to induce a strong guess at the $M_K$ needed to span the gap. Inducing $M_K$ allows the explanation of $I$ to be completed and $I_K$ to be learned.

$O3$. **Take instruction to learn $M_K$, completing the explanation.** The default response of the agent (when the other options are not deemed appropriate) is to ask the instructor to explain $I$ further. The further instruction can teach the agent $M_K$. Again, learning $M_K$ allows the explanation of $I$ to be completed and $I_K$ to be learned.

$O4$. **Abandon the explanation and learn $I_K$ in another way.** The instructional context can indicate that the missing knowledge $M_K$ would be very difficult to learn. This occurs when either the instructor refuses to give further information when asked to, or when the agent has projected multiple operators that may be missing pieces of knowledge (multiple potential $M_K$s). Since it is unknown whether $M_K$ will ever be acquired, the agent abandons its explanation of $I$ altogether. Instead, it attempts to learn $I_K$ directly (using inductive heuristics), without an explanation to base the learning on.

These options will be made clearer through examples presented in the following sections.

Situated explanation is summarized in Table 3. Unlike some knowledge acquisition approaches, it does not include an explicit check for consistency when newly learned knowledge is added to the agent's knowledge base. As Kodratoff and Tecuci (1987a) point out, techniques like situated explanation are biased toward consistency because they only acquire new knowledge when current knowledge is insufficient, and they use current knowledge when deriving new knowledge. However, in some domains, explicit consistency checks (such as those used by Wilkins' (1990) ODYSSEUS) may be required.

Situated explanation meets the requirement that learning be incremental ($T_4$) because it occurs during the ongoing processing of the agent and adds new pieces of knowledge to the agent's memory in a modular way. The local control structure of the PSCM allows new knowledge to be added independent of current knowledge. If there is a conflict between pieces of knowledge (for example, proposing two different operators in the same situation), an impasse will arise that can be reasoned about or resolved with further instruction.

| $T_1$ | General learning from specific cases | |
|---|---|---|
| $T_2$ | Fast learning (each task instructed only once) | |
| $T_3$ | Maximal use of prior knowledge | |
| $T_4$ | Incremental learning | |
| $T_5$ | Knowledge-type flexibility | |
| | a. | state inference |
| | b. | operator proposal |
| | c. | operator control |
| | d. | operator effects |
| | e. | operator termination |
| $I_2$ | Command flexibility | |
| | a. | known command |
| | b. | skipped steps |
| | c. | unknown command |
| $I_3$ | Situation flexibility | |
| | a. | implicitly situated |
| | b. | explicitly situated: hypothetical state |
| | | hypothetical goal |

Table 4: Expanded requirements of tutorial instruction met by INSTRUCTO-SOAR.

## 6. INSTRUCTO-SOAR

INSTRUCTO-SOAR is an instructable agent built within Soar – and thus, the PSCM – that uses situated explanation to learn from tutorial instruction.[7] INSTRUCTO-SOAR engages in an interactive dialogue with its instructor, receiving natural language instructions and learning to perform tasks and extend its knowledge of the domain. This section and the next describe how INSTRUCTO-SOAR meets the targeted requirements of tutorial instruction, which are shown in expanded form in Table 4. This section describes the system's basic performance when learning new procedures, and extending procedures to new situations, from imperative commands (implicitly situated instructions); the next describes learning other types of knowledge and handling explicitly situated instructions.

---

7. For an overview of Soar, and other systems built within it, see (Rosenbloom, Laird, & Newell, 1993b).

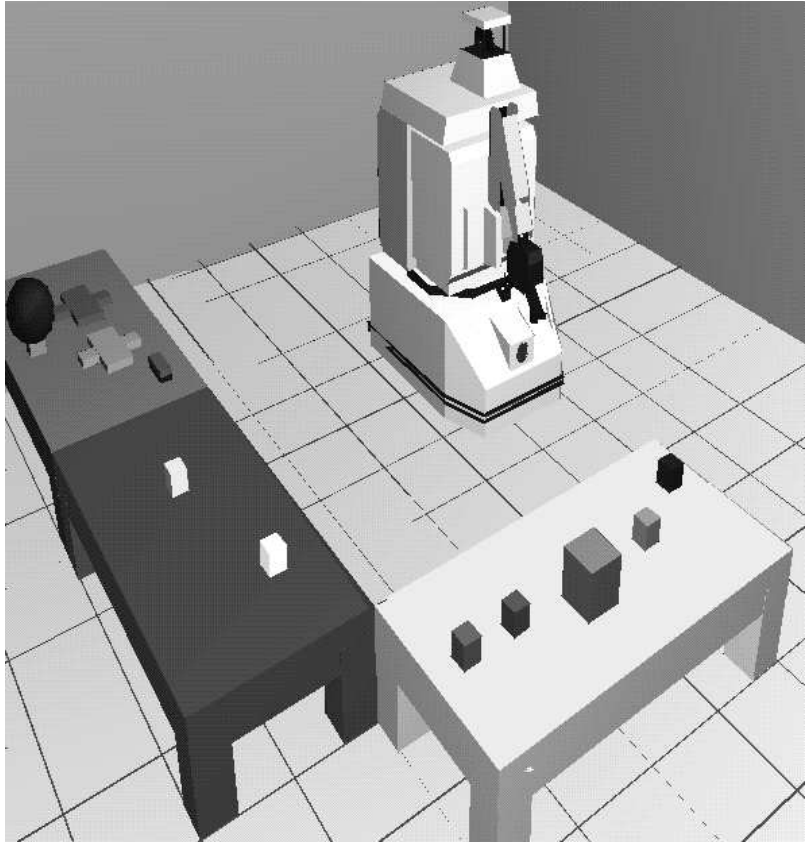

Figure 5: The robotic domain to which INSTRUCTO-SOAR has been applied.

## 6.1 The Domain and the Agent's Initial Knowledge

The primary domain to which INSTRUCTO-SOAR has been applied is the simulated robotic world shown in Figure 5.[8] The agent is a simulated Hero robot, in a room with tables, buttons, blocks of different sizes and materials, an electromagnet, and a light. The magnet is toggled by closing the gripper around it. A red button toggles the light on or off; a green button toggles it dim or bright, when it is on.

INSTRUCTO-SOAR consists of a set of problem spaces within Soar that contain three main categories of knowledge: natural language processing knowledge, originally developed for NL-Soar (Lewis, 1993); knowledge about obtaining and using instruction; and knowledge of the task domain itself. This task knowledge is extended through learning from instruction. INSTRUCTO-SOAR does not expand its natural language capabilities *per se* as it takes instruction, although it does learn how sentences map onto new operators that it learns. It has complete, noiseless perception of its world, and can recognize a set of basic object properties (e.g., `type`, `color`, `size`) and relationships (e.g., robot `docked-at` table,

---

8. The techniques have also been applied in a limited way to a flight domain (Pearson, Huffman, Willis, Laird, & Jones, 1993), in which Soar controls a flight simulator and instructions are given for taking off.

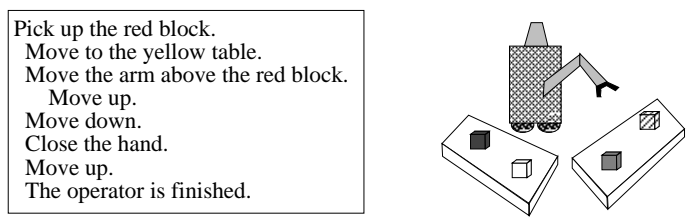

Pick up the red block.
    Move to the yellow table.
    Move the arm above the red block.
        Move up.
    Move down.
    Close the hand.
    Move up.
    The operator is finished.

Figure 6: Instructions given to Instructo-Soar to teach it to pick up a block.

gripper `holding` object, objects `above`, `directly-above`, `left-of`, `right-of` one another). The set of properties and relations can be extended through instruction, as described below.

The agent begins with knowledge of a set of primitive operators to which it can map natural language sentences, and can execute. These include moving to tables, opening and closing the hand, and moving the arm up, down, and above, left of, or right of things. The agent can also internally project these operators. However, some of their effects under various conditions are unknown. For instance, the agent does not know which operators affect the light or magnet, or that the magnet will attract metal objects. Also, the agent begins with no knowledge of complex operators (that involve combinations of primitive operators), such as picking up or arranging objects, pushing buttons, etc.

## 6.2 Learning New Procedures through Delayed Explanation

Instructo-Soar learns new procedures (PSCM operators) from instructions like those shown in Figure 6, for picking up a block. Since "pick up" is not a known procedure initially, when told to "Pick up the red block," the agent realizes that it must learn a new operator.

To perform a PSCM operator, the operator must be *selected*, *implemented*, and *terminated*. To select the operator in the future based on a command requires knowledge of the operator's argument structure (a template), and how natural language maps to this structure. Thus, to learn a new operator, the agent must learn four things:

1. **Template:** Knowledge of the operator's arguments and how they can be instantiated. For picking up blocks, the agent acquires a new operator with a single argument, the object to be picked up.

2. **Mapping from natural language:** A mapping from natural language semantic structures to an instantiation of the new operator, so that the operator can be selected when commanded in the future. For picking up blocks, the agent learns to map the semantic object of "Pick up ..." to the single argument of its new operator template.

3. **Implementation:** How to perform the operator. New operators are performed by executing a sequence of smaller operators. The implementation takes the form of selection knowledge for these sub-operators (e.g., move to the proper table, move the arm, etc.)

4. **Termination conditions:** Knowledge to recognize when the new operator is achieved – the goal concept of the new operator. For "pick up," the termination conditions include holding the desired block, with the arm raised.

Requirement $T_2$ ("fast learning") stipulates that after the first execution of a new procedure, the agent must be able to perform at least the same task without being re-instructed. Thus, the agent must learn, in some form, each of the four parts of a new operator during its first execution.

A general implementation of a new operator can be learned through situated explanation of each of its steps. During the first execution of a new operator, though, the instructions for performing it cannot be explained, because the agent does not yet know the goal of the operator (e.g., the agent does not know the termination conditions of "pick up") or the steps following the current one to reach that goal. However, in this instructional context – explaining instructed steps of a procedure being learned – it is clear that the missing knowledge of the remaining steps and the procedure's goal *will* be acquired later, because the instructor is expected to teach the procedure to completion. Thus, the agent *delays explanation* (option $O1$) and for now memorizes each implementation instruction in a rote, episodic form. At the end of the first execution of a new procedure, the agent induces the procedure's goal – its termination conditions – using a set of simple inductive heuristics. On later executions of the procedure, the original instructions are recalled and explained to learn a general implementation.

We describe the details of this process using the "pick up" example.

### 6.2.1 First Execution

The example, shown in Figure 6, begins with the instruction "Pick up the red block." The agent comprehends this instruction, producing a semantic structure and resolving "the red block" to a block in its environment. However, the semantic structure does not correspond to any known operator, indicating that the agent must learn a new operator (which it calls, say, `new-op14`). To learn a template for the new operator, the agent simply assumes that the argument structure of the command used to request the operator is the required argument structure of the operator itself. In this case, a template for the new operator is generated with an argument structure that directly corresponds to the semantic arguments of the "pick up" command (here, one argument, `object`). The agent learns a mapping from the semantic structure to the new operator's template, to be used when presented with similar requests in the future. This simple approach to learning templates and mappings is sufficient for imperative sentences with direct arguments, but will fail for commands with complex arguments, such as path constraints ("Move the dynamite into the other room, keeping it as far from the heater as possible").

Next, the new operator is selected for execution. Since its implementation is unknown, the agent immediately reaches an impasse and asks for further instructions. Each instruction in Figure 6 is given, comprehended and executed in turn. These instructions provide the implementation for the new operator. They are implicitly situated – each applies to the current situation in which the agent finds itself.

At any point, the agent may be given another command that cannot be directly completed – one that requests either another unknown procedure or a known procedure that the agent does not know how to perform in the current situation due to skipped steps. This is *command flexibility* (requirement $I_2$). For example, within the instructions for "pick up," the command "Move above the red block" cannot be completed because of a skipped step (the arm must be raised to move above something). An impasse arises where the instructor indicates the needed step ("move up"), and then continues instructing "pick up."

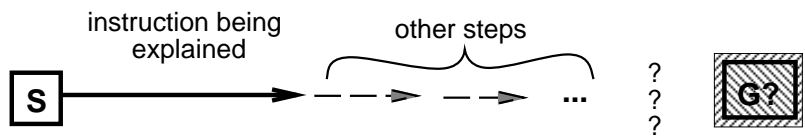

Figure 7: Instructions teaching a new operator cannot be explained before the termination conditions of the new operator are learned.

Ultimately, the implementation of a new operator can be learned at the proper level of generality by explaining each instructed step. However, as illustrated in Figure 7, during its initial execution forming this explanation is impossible, because the goal of the new operator and the other steps (further instructions) needed to reach it are not yet known. Since these missing pieces of the explanation are expected to be available later, the agent delays explanation and resorts to *rote learning* of each instructed step.

In INSTRUCTO-SOAR, rote learning occurs as a side effect of language comprehension. While reading each sentence, the agent learns a set of rules that encode the sentence's semantic features. The rules allow NL-Soar to resolve referents in later sentences, implementing a simple version of Grosz's focus space mechanism (Grosz, 1977). The rules record each instruction, indexed by the goal to which it applies and its place in the instruction sequence. The result is essentially an episodic case that records the specific, lock-step sequence of the instructions given to perform the new operator. For instance, it is recorded that "to pick-up (that is, `new-op14`) the red block, `rb1`, I was first told to move to the yellow table, `yt1`." Of course, the information contained within the case could be generalized, but at this point any generalization would be purely heuristic, because the agent cannot explain the steps of the episode. Thus, INSTRUCTO-SOAR takes the conservative approach of leaving the case in rote form.

Finally, the agent is told "The operator is finished," indicating that the goal of the new operator has been achieved. This instruction triggers the agent to learn termination conditions for the new operator. Learning termination conditions is an inductive concept formation problem: The agent must induce which features of those that hold in the current state imply a positive instance of the new operator's goal being achieved. Standard concept learning approaches may be used here, as long as they produce a strong hypothesis within a small number of examples (due to the "fast learning" requirement, $T_2$). INSTRUCTO-SOAR uses a simple heuristic to strongly bias its induction: It hypothesizes that everything that has *changed* between the initial state when the new operator was requested and the current state is part of the new operator's termination conditions. In this case, the changes are that the robot is docked at a table, holding a block, and the block and gripper are both up in the air.

This heuristic gives a reasonable guess, but is clearly too simple. Conditions that changed may not matter; e.g., perhaps it doesn't matter to picking up blocks that the robot ends up at a table. Unchanged conditions may matter; e.g., if learning to build a "stoplight," block colors are important although they do not change. Thus, the agent presents the induced set of termination conditions to the instructor for possible alteration and verification. The instructor can add or remove conditions. For example, in the "pick up" case the instructor might say "The robot need not be docked at the yellow table" to remove a condition deemed unnecessary, before verifying the termination conditions.

INSTRUCTO-SOAR performs induction by EBL (chunking) over an overgeneral theory that can make inductive leaps (similar to, e.g., Miller, 1993; Rosenbloom & Aasman, 1990; VanLehn, Ball, & Kowalski, 1990). This type of inductive learning has the advantage that the agent can alter the bias to reflect other available knowledge. In this case, the agent uses further instruction (the instructor's indications of features to add or remove) to alter the induction. Other knowledge sources that could be employed (but are not in the current implementation) include analogy to other known operators (e.g., pick up actions in other domains), domain-specific heuristics, etc.

Through the first execution of a new operator, then, the agent:

- Carries out a sequence of instructions achieving a new operator.

- Learns an operator template for the new operator.

- Learns the mapping from natural language to the new operator.

- Learns a rote execution sequence for the new operator.

- Learns the termination conditions of the new operator.

Since the agent has learned all of the necessary parts of an operator, it will be able to perform the same task again without instruction. However, since the implementation of the operator is rote, it can only perform the *exact* same task. It has not learned generally how to pick up things yet.

### 6.2.2 GENERALIZING THE NEW OPERATOR'S IMPLEMENTATION

The agent now knows the goal concept and full (though rote) implementation sequence for the new operator. Thus, it has the information that it needs to explain how each instruction in the implementation sequence leads to goal achievement, provided its underlying domain knowledge is sufficient.

Each instruction is explained by recalling it from episodic memory and internally projecting its effects and the rest of the path to achievement of the termination conditions of the new operator. The projection is a "proof" that the instructed operator will lead to goal achievement in the situation. Soar's chunking mechanism essentially computes the weakest preconditions of the situation and the instruction required for success (similar to standard EBL) to form a general rule proposing the instructed operator. The rule learned from the instruction "Move to the yellow table" is shown in Figure 8. The rule generalizes the original instruction by dropping the table's color, and specializes it by adding the facts that the table has the object sitting on it and that the object is small (only small objects

> **If** the goal is `new-op-14(?obj)`, and
>     `?obj` is `on` table `?t`, and `small(?obj)`, and
>     the robot is not docked at `?t`, and
>     the gripper has `status(open)`,
> **then** propose operator `move-to-table(?t)`.

Figure 8: The general operator proposal rule learned from the instruction "Move to the yellow table" (`new-op-14` is the newly learned "pick up" operator).

can be grasped by the gripper). The rule also tests that the gripper is open, because this condition was important for grasping the block in the instructed case.[9]

After learning general proposal rules for each step in the instruction sequence, the agent can perform the task without reference to the rote case. For instance, if asked to "Pick up the green block," the agent selects `new-op14`, instantiated with the green block. Then, general sub-operator proposal rules like the one in Figure 8 fire one by one, as they match the current situation, to implement the operator. After performing all of the implementation steps, the agent recognizes that the termination conditions are met (the gripper is raised and holding the green block), and `new-op14` is terminated.

Since the general proposal rules for implementing the task are directly conditional on the state, the agent can perform the task starting from any state along the implementation path and can react to unexpected conditions (e.g., another robot stealing the block). In contrast, the rote implementation that was initially learned applied only when starting from the original starting state, and was not reactive because its steps were not conditional on the current state.

## 6.3 Recall Strategies

We have described how our agent recalls and explains each step of a new operator's implementation sequence, after the operator's termination conditions are induced. There are still two open issues: (A) At what point after learning the termination conditions should the agent perform the recall/projection?, and (B) How many steps should be recalled and projected in sequence at a time?

To investigate these issues, we have implemented two different recall/project strategies:

1. **Immediate/complete recall.** The agent recalls and attempts to explain the *full* sequence of instructions for the new operator *immediately* after learning the new operator's termination conditions.

2. **Lazy/single-step recall.** The agent recalls and attempts to explain *single* instructions in the sequence *when asked to perform the operator again* starting from the same initial state. That is, at each point in the execution of the operator, the agent recalls the next instruction, and attempts to explain it by forward projecting it. However, if this projection does not result in a path to goal achievement without any further instructions being recalled, then rather than recalling the next instruction in the sequence to continue the forward projection, the agent gives up on explaining this instruction and simply executes it in the external world.

---

9. More technical details of how Soar's chunking mechanism forms such rules can be found in (Huffman, 1994; Laird, Congdon, Altmann, & Doorenbos, 1993).

These strategies represent the extremes of a continuum of strategies.[10] The strategy to use is a parameter of the agent; it does not dynamically select between strategies while it is running. A possible extension would be to reason about the time pressure in different situations to select the appropriate strategy. Next, we briefly describe the implications of each recall strategy.

### 6.3.1 IMMEDIATE/COMPLETE RECALL STRATEGY

Immediate/complete recall and explanation involves internally projecting multiple operators (the full instruction sequence) immediately after the first execution of the new operator. The projection begins at the state the agent was in when the new operator was first suggested. If the projection successfully achieves the termination conditions of the new operator, the agent learns general implementation rules for every step. The advantage of this strategy is that the agent learns a general implementation for the new operator immediately after its first execution (e.g., the agent can pick up other objects right away).

The strategy has three important disadvantages. First, it requires that the agent reconstruct the initial state in which it was commanded to perform the new operator. This reconstruction may be difficult if the amount of information in the state is large (although it is not in the small robotic domain being used here).

Second, recall and projection of the entire sequence of instructed steps is time-consuming, requiring time proportional to the length of the instruction sequence. During the process, the agent's performance of tasks at hand is suspended. This suspension could be awkward if the agent is under pressure to act quickly.

Third, as illustrated in Figure 9, multiple step projections are susceptible to compounding of errors in underlying domain knowledge. The projection of each successive operator begins from a state that reflects the agent's knowledge of the effects of prior operators in the sequence. If this knowledge is incomplete or incorrect, the state will move further and further from reflecting the actual effects of prior operators. Minor domain knowledge problems in the knowledge of individual operators, that alone would not produce an error in a single step explanation, may combine within the projection to cause an error. This can lead to incomplete explanations or (more rarely) to spuriously successful explanations (e.g., reaching success too early in the instruction sequence).

### 6.3.2 LAZY/SINGLE-STEP RECALL STRATEGY

In the lazy/single-step recall strategy, the agent waits to recall and explain instructions until asked to perform the new operator a second time from the same initial state. In addition, the agent only recalls a single instruction to internally project at a time. After the recalled operator is projected, the agent applies whatever general knowledge it has about the rest of the implementation of the new operator. This general knowledge, however, does not include rote memories of other past instructions. That is, if the agent does not know the rest of the path to complete the new operator using general knowledge, it does not recall any further instructions in the sequence from its rote memories. Rather, the internal projection is terminated and the single recalled operator is applied in the external world.

---

10. We also implemented a lazy/complete recall strategy, which will not be described here (see Huffman, 1994, for details).

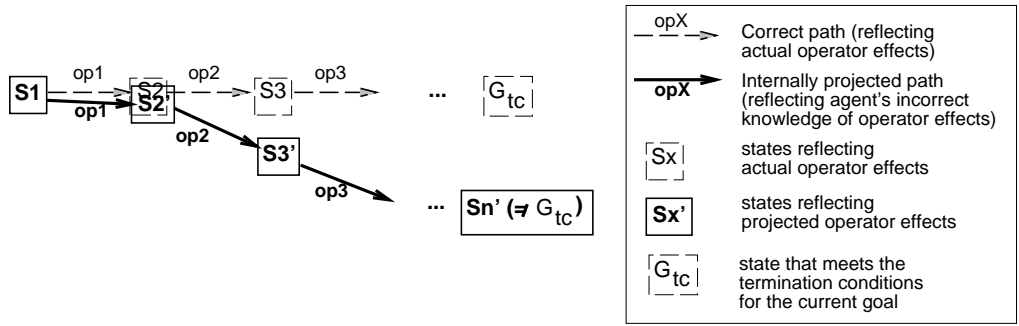

Figure 9: Multiple step projections can result in incomplete explanations due to compounding of errors in domain knowledge.

This strategy addresses the three disadvantages of the immediate/complete strategy. First, it does not require reconstruction of the original instruction state; rather, it waits for a similar state to occur again.

Second, recalling and projecting a single instruction at a time does not require a time-consuming introspection that suspends the agent's ongoing activity. For "pick up," for instance, Table 5 shows the longest time that the agent's external action (movements or instruction requests) is suspended using each strategy (as measured in Soar decision cycles, which last about 35 milliseconds each for INSTRUCTO-SOAR on an SGI R4400 Indigo). The immediate/complete strategy does no external actions for 304 decision cycles (about 11 seconds on our Indigo) immediately following the first execution, in order to recall and explain the complete instruction sequence. Using the lazy/single-step strategy, only one instruction is ever recalled/explained at a time before action is taken in the world; thus, the longest time without action is only 75 decision cycles (about 2 seconds). The total recall/explanation time is proportional to the length of the instruction sequence in both cases (304 vs. 294 decision cycles), but in the lazy/single-step strategy, that time is interleaved with the execution of the instructions rather than fully taken after the first execution.

Third, the lazy/single-step strategy overcomes the problem of compounding of domain theory errors by beginning the projection of each instruction from the current state of the world after external execution of the previous instructions. Thus, the beginning state of each projection correctly reflects the effects of the previous operators in the implementation sequence.

The major disadvantage of this strategy is that it requires a number of executions of the new operator equal to the length of the instruction sequence in order to learn the whole general implementation. This is because limiting recall to a single step allows only a single sub-operator per execution to be generalized. This disadvantage, however, leads to two interesting learning characteristics:

- **Back-to-front generalization**. Generalized learning starts at the end of the implementation sequence and moves towards the beginning. On the second execution of the new operator, a path to the goal is known only for the last instruction in the sequence (it leads directly to goal completion), so a general proposal for that instruction is learned. On the third execution, after the second to last instruction is projected, the proposal learned previously for the last operator applies, leading to goal achievement and allowing a general proposal for the second to last instruction to be learned. This pattern continues back through the entire sequence until the full implementation is learned generally. As Figure 10 shows, the resulting learning curves closely approximate the power law of practice (Rosenbloom & Newell, 1986) ($r = 0.98$ for both (a) and (b)).

- **Effectiveness of hierarchical instruction**. Due to the back-to-front effect, the agent learns a new procedure more quickly when its steps are taught using a hierarchical organization than when they are taught as a flat sequence. Figure 11 depicts a flat, nine-step instruction sequence for teaching INSTRUCTO-SOAR to move one block left of another; Figure 12 depicts a hierarchical instruction sequence for the same procedure, that contains 13 instructed steps, but a maximum of 3 in any subsequence. By breaking the instruction sequence into shorter subsequences, a hierarchical organization allows multiple subtrees of the hierarchy to be generalized during each execution. General learning for an $N$ step operator takes $N$ executions using a flat instruction sequence. Taught hierarchically as an $H$-level hierarchy with $\sqrt[H]{N}$ subtasks in each subsequence, only $H \cdot \sqrt[H]{N}$ executions are required for full generalization. The hierarchy in Figure 12 has an irregular structure, but results in a speedup because the length of every subsequence is small (in this case, smaller than $\sqrt{N}$). Empirically, the flat sequence of Figure 11 takes nine ($N$) executions to generalize, whereas the hierarchical sequence takes only six. Hierarchical organization has the additional advantage that more operators are learned that can be used in future instructions.

| | Immediate/complete | Lazy/single-step |
|---|---|---|
| Largest time without external action | 304 | 75 |
| Largest total recall/explanation time during an execution | 304 (end of 1st exec'n) | 294 (during 2nd exec'n.) |

Table 5: Timing comparison, in Soar decision cycles, for learning "pick up" using the immediate/complete and lazy/single-step recall strategies.

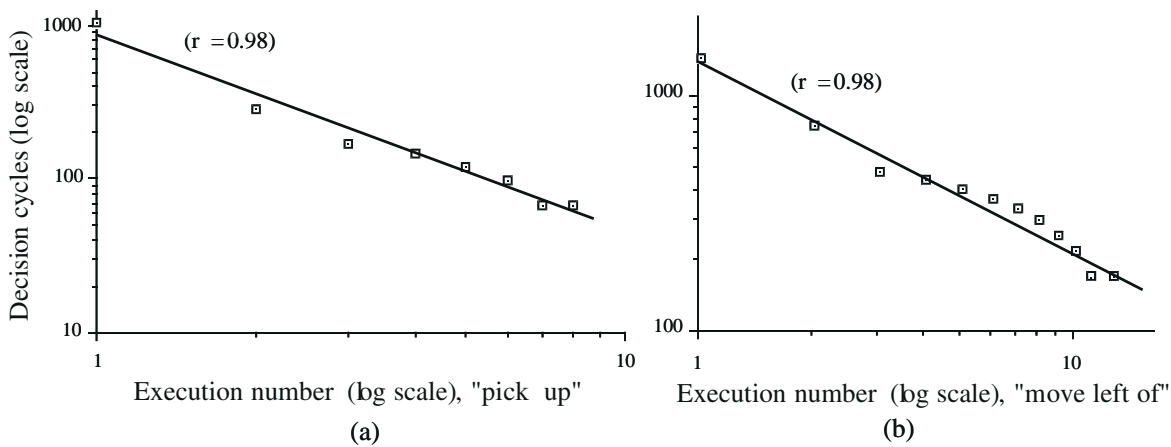

(a)  (b)

Figure 10: Decision cycles versus execution number to learn to (a) pick up and (b) move objects left of one another, using the lazy/single-step strategy.

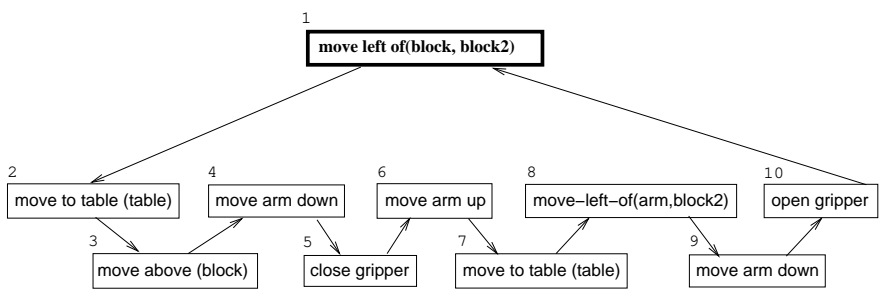

Figure 11: A graphical view of a flat instruction sequence for `move-left-of(block,block2)`.

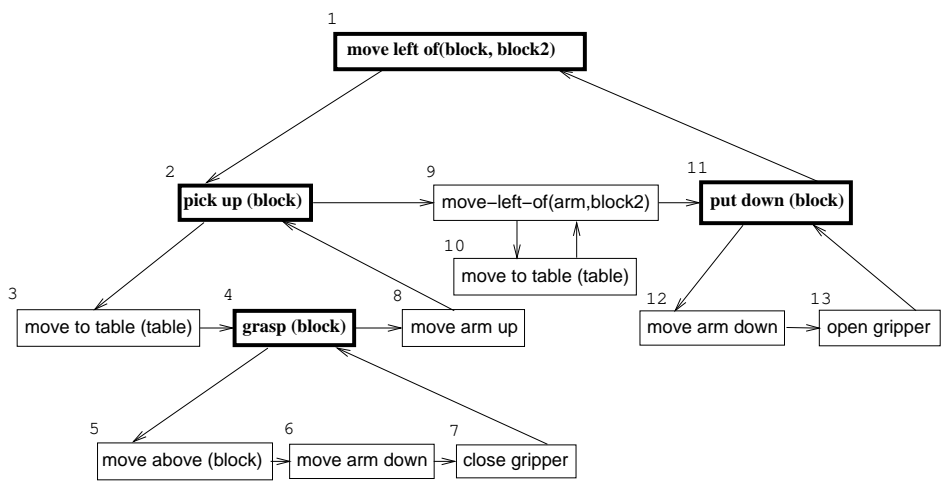

Figure 12: A graphical view of a hierarchical instruction sequence for `move-left-of(block,block2)`. New operators are shown in bold.

## 6.4 Supporting Command Flexibility

Command flexibility (requirement $I_2$) stipulates that the instructor may request either an unknown procedure, or a known procedure that the agent does not know how to perform in the current state (skipping steps), at any point. This can lead to multiple levels of embedded instruction. As we have seen, Instructo-Soar learns completely new procedures from instructions for unknown commands. In addition, when the agent is asked to perform a known procedure in an unfamiliar situation – one from which the agent does not know what step to take – it learns to extend its knowledge of the procedure to that situation.

An example is contained in the instructions for "Pick up the red block," when the agent is asked to "Move above the red block." The agent knows how to perform the operator when its arm is raised. However, in this case the arm is lowered, and so the agent reaches an impasse and asks for further instruction.[11] When told to "Move up," the agent internally projects raising its arm, which allows it to achieve moving above the red block. From this projection it learns the general rule: move the arm up when trying to move above an object that is on the table the agent is docked at. This rule extends the "move above" procedure to cover this situation.

Any operator – even one previously learned from instruction – may require extension to apply to a new situation. This is because when the agent learns the general implementation for a new operator, it does not reason about all *possible* situations in which the operator might be performed, but limits its explanations to the series of situations that arises during the actual execution of the new operator while it is being learned.

Newly learned operators may be included in the instructions for later operators, leading to learning of operator hierarchies. One hierarchy of operators learned by Instructo-Soar is shown in Figure 13. Learning procedural hierarchies has been identified as a fundamental component of children's skill acquisition from tutorial instruction (Wood, Bruner, & Ross, 1976). In learning the hierarchy of Figure 13, Instructo-Soar learned four new operators, an extension of a known operator (`move above`), and an extension of a new operator (extending "pick up" to work if the robot already is holding a block). Because of command flexibility, this same hierarchy can be taught in exponentially many different ways (Huffman, 1994). For instance, new operators that appear as sub-operators (e.g., `grasp`) can be taught either before or during teaching of higher operators (e.g., `pick up`).

## 6.5 Abandoning Explanation when Domain Knowledge is Incomplete

All of the general operator implementation learning described thus far depends on explaining instructions using prior domain knowledge (as opposed to the learning of operator termination conditions, which is inductive). What if the domain knowledge is incomplete, making explanation impossible? For sequences of multiple operators, pinpointing what knowledge is missing is an extremely difficult credit assignment problem (sequences known to contain only one operator, however, are a more constrained case, as described in the next section).

---

11. Another option would be to search; i.e., to apply a weak method such as means-ends analysis. In this example, the search would be easy; in other cases, it could be costly. In any event, since the goal of Instructo-Soar is to investigate the use of instruction, our agent always asks for instructions when it reaches an impasse in task performance. Nothing in Instructo-Soar precludes the use of search or knowledge from other sources, however.

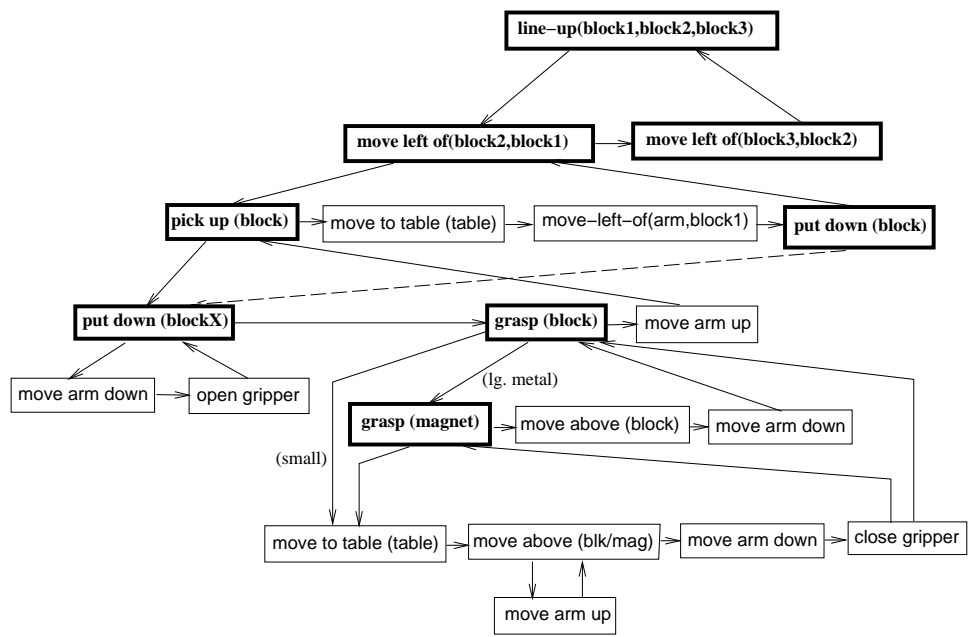

Figure 13: A hierarchy of operators learned by INSTRUCTO-SOAR. Primitive operators are in light print; learned operators are in bold.

In general, an explanation failure that is detected at the end of the projection of an instruction sequence could be caused by missing knowledge about any operator in the sequence. Thus, when faced with an incomplete explanation of a sequence of multiple instructions, INSTRUCTO-SOAR abandons the explanation and instead tries to induce knowledge directly from the instructions (option $O4$).

As an example, consider a case in which all of INSTRUCTO-SOAR's knowledge of secondary operator effects (frame axiom type knowledge) is removed before teaching it a procedure. For example, although the agent knows that closing the hand causes it to have status closed, it no longer knows that closing the hand around a block causes the block to be held. Now, the agent is taught a new procedure, such as to pick up the red block. After the first execution, the agent attempts to recall and explain the instructions as usual, but fails because of the missing knowledge. That is, the block is not picked up during the projection of the instructions, since the agent's knowledge does not indicate that it is held. The agent records the fact that this procedure's instructions cannot be explained.

Later, the agent is again asked to perform the procedure, and again recalls the instructions. However, it also recalls that explaining the instructions failed in the past. Thus, it abandons explanation and instead attempts to induce a general proposal rule directly from each instruction.[12]

---

12. Since an incomplete explanation for a procedure may indicate that some effect(s) of an operator in the instruction sequence is unknown, another alternative (not yet implemented in INSTRUCTO-SOAR) would be for the agent to observe the effects of each operator in the sequence as it is performed, comparing the observations to the effects predicted by domain knowledge. Any differences would allow the agent

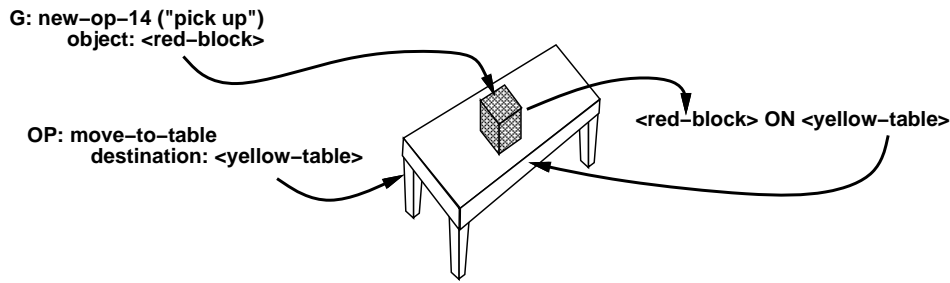

Figure 14: The use of the **OP-to-G-path** heuristic, with $OP$ "move to the yellow table," and $G$ "pick up the red block."

In the "pick up" example, the agent first recalls the command to move to the yellow table. To learn a proposal rule for this operator (call it $OP$), the agent must induce a set of conditions of the state under which performing $OP$ will contribute to achieving the "pick up" goal (call it $G$). Instructo-Soar uses two simple heuristics to induce these state conditions:

- **OP-to-G-path.** For each object $Obj1$ filling a slot of $OP$, and each object $Obj2$ attached to $G$, include the shortest existing path (heuristically of length less than three) of relationships between $Obj1$ and $Obj2$ in the set of induced conditions.

  This heuristic captures the intuition that if an operator involves some object, its relationship to the objects relevant to the goal is probably important. Figure 14 shows its operation for "move to the yellow table." As the figure indicates, there is a path between $G$'s object, the red block, and the destination of $OP$, the yellow table, through the relationship that the block is **on** the table.

- **OP-features-unachieved.** Each termination condition (essentially, each primary effect) of $OP$ that is *not achieved* in the state before $OP$ is performed is considered an important condition.

  This heuristic captures the intuition that all of the primary effects of $OP$ are probably important; therefore, it matters that they are not achieved when $OP$ is selected. In our example, $OP$'s primary effect is that the robot ends up docked at the table; thus, the fact that the robot is *not* initially docked at the table is added to the inferred set of conditions for proposing $OP$.

These heuristics are implemented as Soar operators that compute the appropriate conditions. Once a set of conditions is induced, it is presented to the instructor, who can add or remove conditions before verifying them. Upon verification, a rule is learned proposing $OP$ (e.g., `move-to-table(?t)`) when the induced conditions hold (e.g., `goal is pick-up(?b)`, `?b isa block, on(?b,?t)`). This rule is similar to the rule learned from explanation (Figure 8), but only applies to picking up a `block` (overspecific), and does not stipulate that the

---

to learn new operator effects that could complete the explanation of the procedure. Learning effects of operators from observation has been explored by a number of researchers (Carbonell & Gil, 1987; Pazzani, 1991b; Shen, 1993; Sutton & Pinette, 1985; Thrun & Mitchell, 1993).

object must be `small` (overgeneral). A similar induction occurs for each step of "pick up," so that the agent learns a general implementation for the full "pick up" operator. However, unless corrections are made by the instructor, this induced implementation is not as correct as one learned from explanation; for instance, it applies (wrongly) to any block instead of to any small object. In a more complex domain, inferring implementation rules would be even less successful. Not surprisingly, psychological research shows that human subjects' learning from procedural instructions also degrades if they lack domain knowledge (Kieras & Bovair, 1984).

Returning to the targeted instruction requirements in Table 4, INSTRUCTO-SOAR's learning of procedures illustrates ($T_1$) general learning from specific instructions, ($T_2$) fast learning (because each procedure need only be instructed once) by ($T_3$) using prior domain knowledge to construct explanations, and ($T_4$) incremental learning during the agent's ongoing performance. Two types of PSCM knowledge are learned: ($T_5$(b)) operator proposals for sub-operators of the procedure, and ($T_5$(e)) the procedure's termination conditions. The learning involves either delayed explanation, or when domain knowledge is inadequate, abandoning explanation in favor of simple induction. The instructions are each ($I_3$(a)) implicitly situated imperative commands, for either ($I_2$(a)) known procedures, ($I_2$(b)) known procedures where steps have been skipped, or ($I_2$(c)) unknown procedures.

## 7. Beyond Imperative Commands

Next, we turn to learning the remaining types of PSCM knowledge ($T_5$(a,c,d)) from various kinds of explicitly situated instructions ($I_3$(b)). From an explicitly situated instruction, INSTRUCTO-SOAR constructs a *hypothetical* situation (goal and state) that includes the objects, properties, and relationships mentioned explicitly in the instruction as well as any features of the current situation that are needed to carry out the instruction.[13] This hypothetical situation is used as the context for a situated explanation of the instruction.

### 7.1 Hypothetical Goals and Learning Effects of Operators

A goal is explicitly specified in an instruction by a *purpose clause* (DiEugenio, 1993): "*To do X*, do Y." The basic knowledge to be learned from such an instruction is an operator proposal rule for doing Y when the goal is to achieve X.

Consider this example from INSTRUCTO-SOAR's domain:

> `> To turn on the light, push the red button.`

The agent has been taught how to push buttons, but does not know the red button's effect on the light. From a purpose clause instruction like this example, the agent creates a hypothetical situation with the goal stated in the purpose clause (here, "turn on the light"), and a state like the current state, but with that goal not achieved (here, with the light off). Within this situation, the agent attempts to explain the instruction by forward projecting the action of pushing the red button.

If the agent knew that pushing the red button toggles the light, then in the projection, the light would come on. Thus, the explanation would succeed, and a general operator proposal rule would be learned, that proposed pushing the red button when the light is off and the goal is to turn it on.

---

13. See (Huffman, 1994) for details of how these features are determined.

However, since in actuality the agent is missing the knowledge ($M_K$) that pushing the button affects the light, the light does not come on within the projection. The explanation is incomplete.

When Instructo-Soar's explanation of a *sequence* of operators fails, the agent does not try to induce the missing knowledge needed to complete the explanation, because it could be associated with any of the multiple operators. Rather, the explanation is simply abandoned, as described in Section 6.5. However, in this case, the unexplainable sequence contains only *one* operator. In addition, the form of the instruction gives the agent a strong expectation about that operator's intended effect. Based on the purpose clause, the agent expects that the specified action (pushing the button) will cause achievement of the specified goal (turning on the light). DiEugenio (1993) found empirically that this type of expectation holds for 95% of naturally occurring purpose clauses.

The expectation constrains the "gap" in the incomplete explanation: the state after pushing the button *should* be a state with the light on, and only one action was performed to produce this effect. Based on this constrained gap, the agent attempts to induce the missing knowledge $M_K$ in order to complete the explanation (option $O2$). The most straightforward inference of $M_K$ is simply that an unknown effect of the single action is to produce the expected goal conditions – e.g., pushing the button should cause the light to come on. The instructor is asked to verify this inference.[14]

Once it is verified, Instructo-Soar heuristically guesses at the state conditions under which the effect will occur. It uses the **OP-to-G-path** heuristic as a very naive causality theory (Pazzani, 1991a) to guess at the causes of the inferred operator effect. Here, **OP-to-G-path** notices that the light and the red button are both on the same table. In addition, the agent includes the fact that the inferred effect did not hold (the light was off) before the operator caused it. The result is presented to the instructor:

> *I think that doing push the button causes:*
> > *the light to be on*
> *under the following conditions:*
> > *the light is not currently on, the light is on the table, the button is on the table*
> *Are those the right conditions?*

Here, the heuristics have not recognized that it matters which button is pushed (the red one). The instructor can add this condition by saying ``The button must be red.'' Once the instructor verifies the conditions, the agent adds the new piece of operator effect knowledge to its memory:

```
if   projecting push-button(?b), and
     ?l isa light with status off, on table ?t, and
     ?b isa button with color red, on table ?t,
then light ?l now has status on.
```

---

14. If the inference is rejected, the agent abandons the explanation and directly induces a proposal rule for pushing the button from the instruction, as described in Section 6.5.

Immediately after being learned, this rule applies to the light in the forward projection for the current instruction. The light comes on, completing the instruction's explanation by achieving its goal. From this explanation, the agent learns the proposal rule that proposes pushing the red button when the goal is to turn on the light. Thus, the agent has acquired new knowledge at multiple levels; inferring an unknown effect of an operator supported learning a proposal for that operator.

This example illustrates ($I_3$(b)) the use of hypothetical goal instructions and the use of option $O2$ for dealing with incomplete explanations – inferring missing knowledge – to learn new operator effects ($T_5$(d)), thus extending domain knowledge.

## 7.2 Hypothetical States to Learn About Contingencies

Instructors use instructions with hypothetical states (e.g., conditionals: "If [state conditions], do ...") either to teach general policies ("If the lights are on when you leave the room, turn them off.") or to teach contingencies when performing a task. INSTRUCTO-SOAR handles both of these; here, we will describe the latter.

A contingency instruction indicates a course of action to be followed when the current task is performed in a future situation different from the current situation. Instructors often use contingency instructions to teach about situations that differ from the current one in some crucial way that should alter the agent's behavior. Contingency instructions are very common in human instruction; Ford and Thompson (1986) found that 79% of the conditional statements in an instruction manual communicated contingency options to the student.

Consider this interaction:

> `> Grasp the blue block.`
> *That's a new one for me. How do I do that?*
>     `> If the blue block is metal, then pick up the magnet.`

The blue block is not made of metal, but the instructor is communicating that if it *were*, a different course of action would be required.

From the conditional instruction "If the blue block is metal, then pick up the magnet," the agent needs to learn an operator proposal rule for picking up the magnet under appropriate conditions. The agent begins by constructing the hypothetical situation to which "pick up the magnet" applies. "If the blue block is metal" indicates a hypothetical state that is a variant of the current state with the blue block having `material metal`. The current goal ("Grasp the blue block") is also the goal in the hypothetical situation.

Within this situation, the agent projects picking up the magnet to explain how it will allow the block to be grasped. However, the agent is missing much of the knowledge needed to complete this explanation. It does not know the goal concept of "Grasp" yet, or the rest of the instructions to reach that goal.

Since the instruction being explained is for a contingency, the rest of the instructions that the agent is given to "Grasp the blue block" may not (and in this case, do not) apply to the contingent situation, where the block is metal. In the normal grasp sequence, for instance, the agent learns to close its hand around the grasped object, but when grasping a metal object, the hand is closed around the magnet. Since knowledge of how to complete grasping a metal object is needed to explain the contingency instruction, and the agent does not know when it might learn this missing knowledge, it abandons the explanation (option $O4$). Instead, it uses the heuristics described in Section 6.5 to directly induce an operator proposal rule for "Grasp the magnet." In addition to the conditions generated by the heuristics, the conditions indicated in the antecedent of the instruction are included. The result is presented to the instructor for alteration and verification:

> *So I'm guessing the conditions for doing "pick up the magnet"*
> *when your goal is "grasp the block" are:*
> *the block is metal*
> *Is that right?*
> `> Right.`

From this interaction the agent learns a rule that proposes picking up the magnet when the goal is to grasp a metal block. After this learning is completed, since the agent has not yet finished grasping the blue block, it continues to receive instruction for that task. Further contingencies can be indicated at any point. Learning contingencies illustrates ($I_3$(b)) the handling of hypothetical state instructions.

## 7.3 Learning to Reject Operators

Our final examples illustrate learning to reject an operator – a type of operator control knowledge in the PSCM. The examples also detail the remaining option for dealing with incomplete explanations: ($O3$) completing an explanation through further instruction.

Consider these instructions:

> `> Never grasp green blocks.`
> *Why?*
> (a) `> Trust me.`
> (b) `> Green blocks are explosive.`

A negative imperative prohibits a step from applying to a hypothetical situation in which it might apply. Thus, INSTRUCTO-SOAR creates a hypothetical situation in which the prohibited action might be executed; in this case, a state with a graspable green block. Since no goal is specified by the instruction, and there is no other current goal, a default goal of "maintaining happiness" (which is always considered one of the agent's current goals) is used. From this hypothetical situation, the agent internally projects the "grasp" action, expecting an "unhappy" result. However, the resulting state, in which the agent is grasping a green block, is acceptable according to the agent's knowledge. Thus, the projection does not explain why the action is prohibited.

The agent deals with the incomplete explanation by asking for further instruction, in an attempt to learn $M_K$ and complete the explanation. However, the instructor can decline to give further information by saying (a) `Trust me`. Although the instructor will not provide $M_K$, because the prohibition of a single operator (grasping the green block) is being explained, the agent can induce a plausible $M_K$ that will complete the explanation (option $O2$). Since the agent knows that the final state after the prohibited operator is meant to be "unhappy", it simply induces that this state is to be avoided. This is the converse of learning to recognize when a desired goal has been reached (learning an operator's termination conditions). The agent conservatively guesses that *all* of the features of the hypothetical state (here, that there is a green block that is held), taken together, make it a state to be avoided. Because this inference is so conservative, in the current implementation the instructor is not even asked to verify it. The state inference rule that results is as follows:

```
if   goal is ''happiness'', and
     ?b isa block with color green, and
     holding(gripper,?b),
then this state fails to achieve ''happiness''.
```

This rule applies to the final state in the projection of "Never grasp..." The state's failure to achieve happiness completes the agent's explanation of why it should "Never grasp...," and it learns a rule that rejects any proposed operator for grasping a green block.

Alternatively, the instructor could provide further instruction, as in (b) `Green blocks are explosive`. Such instruction can provide the missing knowledge $M_K$ needed to complete an incomplete explanation (option $O3$). From (b), the agent learns a state inference rule: `blocks` with `color green` have `explosiveness high`. INSTRUCTO-SOAR learns state inferences from simple statements like (b), and from conditionals (e.g., "If the magnet is powered and directly above a metal block, then the magnet is stuck to the block") by essentially translating the utterance directly into a rule.[15] Such state inference instructions can be used to introduce new features that extend the agent's representation vocabulary (e.g, `stuck-to`).

The rule learned from "Green blocks are explosive" adds `explosiveness high` to the block that the agent had simulated grasping in the hypothetical situation. The agent knows that touching an explosive object may cause an explosion – a negative result. This negative result completes the explanation of "Never grasp...," and from it the agent learns to avoid grasping objects with `explosiveness high`.

Completing an explanation through further instruction (as in (b)) can produce more general learning than heuristically inferring missing knowledge (as in (a)). In (b), if the agent is later told `Blue blocks are explosive`, it will avoid grasping them as well. In general, multiple levels of instruction can lead to higher quality learning than a single level because learning is based on an explanation composed from strong lower-level knowledge ($M_K$) rather than inductive heuristics alone. $M_K$ (here, the state inference rule) is also available for future use.

Because the agent has learned not only to reject the "grasp" operator but to recognize the bad state that performing it would lead to, the agent can recognize the bad state if it is reached from another path. For instance, the agent can be led through the individual steps of grasping an explosive block without the instructor ever mentioning "grasp." When the agent is finally asked to "Close the gripper" around the explosive object, it does so, but then immediately recognizes the undesirable state it has arrived in and reverses the `close-gripper` action. In the process, it learns to reject `close-gripper` if the hand is around an explosive object, so that in the future it will not reach the undesirable state through this path.

---

15. This translation occurs by chunking, but in an uninteresting way. INSTRUCTO-SOAR does not use explanation to learn state inferences. An extension would be to try to explain why an inference holds using a deeper causal theory.

Notice here the effect of the *situated* nature of INSTRUCTO-SOAR's learning. The agent learns to avoid operators that lead to a bad state only when they *arise* in the agent's performance. Its initial learning about the bad state is recognitional rather than predictive. Alternatively, when the agent first learns about a bad state, it could do extensive reasoning to determine every possible operator that could lead to that state, from every possible previous state, to learn to reject those operators at the appropriate times. This unsituated reasoning would be very expensive; the agent would have to reason through a huge number of possible situations. In addition, whenever new operators were learned, the agent would have to reason about all the possible situations in which they could arise, to learn if they could ever led to a bad state. Rather than this costly reasoning, INSTRUCTO-SOAR simply learns what it can from its situations as they arise.

Another alternative for completely avoiding bad states would be to think through the effects of every action before taking it, to see if a bad state will result. This highly cautious execution strategy would be appropriate in dangerous situations, but is not appropriate in safer situations where the agent is under time pressure. (Moving between more or less cautious execution strategies is not currently implemented in INSTRUCTO-SOAR.)

The "Never grasp..." examples have illustrated the agent's learning of one type of operator control knowledge, namely operator rejection ($T_5$(c)), learning of state inferences ($T_5$(a)), and the use of further instruction to complete incomplete explanations (option $O3$). The final category of learning we will discuss is a second type of operator control knowledge.

## 7.4 Learning Operator Comparison Knowledge

Another type of control knowledge besides operator rejection rules is operator comparison rules, which compare two operators and express a preference for one over the other in a given situation. INSTRUCTO-SOAR learns operator comparison rules by asking for the instructor's feedback when multiple operators are proposed at the same point to achieve a particular goal. Multiple operators can be proposed, for instance, when the agent has been taught two different methods for achieving the same goal (e.g., to pick up a metal block either using the magnet or directly with the gripper). The instructor is asked to either select one of the proposed operators or to indicate that some other action is appropriate. Selecting one of the proposed choices causes the agent to learn a rule that prefers the selected operator over the other proposed operators in situations like the current situation. Alternatively, if the instructor indicates some other operator outside of the set of proposed operators, INSTRUCTO-SOAR attempts to explain that operator in the usual way, to learn a general rule proposing it. In addition, the agent learns rules preferring the instructed operator to each of the other currently proposed operators.

There are two weaknesses to INSTRUCTO-SOAR's learning of operator comparison rules. First, the instructor can be required to indicate a preference for each step needed to complete a procedure, rather than simply choosing between overall methods. That is, the instructor cannot say "Use the method where you grab the block with your gripper, instead of using the magnet," but must indicate a preference for each individual step of the method employing the gripper. This is because in the PSCM, knowledge about steps in a procedure is accessed independently, as separate proposal rules, rather than as an aggregate method. Independent access improves flexibility and reactivity – the agent can combine steps from different methods as needed based on the current situation – but a higher level grouping of steps would simplify instruction for selecting between complete methods.

The second weakness is that although the agent uses situated explanation to explain the selection the instructor makes, it does not explain why that selection is *better* than the other possibilities. Preferences between viable operators are often based on global considerations; e.g., "Prefer actions that lead to overall faster/cheaper goal achievement." Learning based on this type of global preference (which in turn may be learned through instruction) is a point for further research.

## 8. Discussion of Results

We have shown how INSTRUCTO-SOAR learns from various kinds of instructions. Although the domain used to demonstrate this behavior is simple, it has enough complexity to exhibit a variety of the different types of instructional interactions that occur in tutorial instruction.

Of the 11 requirements that tutorial instruction places on an instructable agent (listed in Table 1), INSTRUCTO-SOAR meets 7 (listed in expanded form in Table 4) either fully or partially. Three of these in particular distinguish INSTRUCTO-SOAR from previous instructable systems:

- **Command flexibility:** The instructor can give a command for any task at each instruction point, whether or not the agent knows the task or how to perform it in the current situation.

- **Situation flexibility:** The agent can learn from both implicitly situated instructions and explicitly situated instructions specifying either hypothetical goals or states.

- **Knowledge-type flexibility:** The agent is able to learn each of the types of knowledge it uses in task performance (the five PSCM types) from instruction.

Earlier, we claimed that handling tutorial instruction's flexibility requires a breadth of learning and interaction capabilities. Combining command, situation, and knowledge-type flexibility, INSTRUCTO-SOAR displays 18 distinct instructional capabilities, as listed in Table 6. This variety of instructional behavior does not require 18 different learning techniques, but arises as one general technique, situated explanation in a PSCM-based agent, is applied in a range of instructional situations.

Our series of examples has illustrated how situated explanation uses an instruction's situation and context during the learning process. First, the situation to which an instruction applies provides the endpoints for attempting to explain the instruction. Second, the instructional context can indicate which option to follow when an explanation cannot be completed. The context of learning a new procedure indicates that delaying explanation (option $O1$) is best, since the full procedure will eventually be taught. If a step cannot be explained in a previously taught procedure, missing knowledge could be anywhere in the procedure, so it is best to abandon explanation (option $O4$) and learn another way. Instructions that provide an explicit context, such as through a purpose clause, localize missing knowledge by giving strong expectations about a single operator that should achieve a single goal. This localization makes it plausible to induce missing knowledge and complete the explanation (option $O2$). In other cases, the default is to ask for instruction about missing knowledge to complete the explanation (option $O3$).

| | Instructional capability | Example |
|---|---|---|
| 1. | Learning completely new procedures | `pick up` |
| 2. | Extending a procedure to apply in a new situation | move up to move above |
| 3. | Hierarchical instruction: handling instructions for a procedure embedded in instruction for others | teaching `pick up` within `line up` |
| 4. | Altering induced knowledge based on further instruction | removing `docked-at` from `pick up`'s termination conditions |
| 5. | Learning procedures inductively when domain knowledge is incomplete | learning with secondary operator effects knowledge removed |
| 6. | Learning to avoid prohibited actions | "Never grasp red blocks." |
| 7. | More general learning due to further instruction | Avoid grasping because "Red blocks are explosive." |
| 8. | Learning to avoid indirect achievement of a bad state | closing hand around explosive block |
| 9. | Inferences from simple specific statements | "The grey block is metal." |
| 10. | Inferences from simple generic statements | "White magnets are powered." |
| 11. | Inferences from conditionals | "if *condition* [and *condition*]* then *concluded state feature*" |
| 12. | Learning an operator to perform for a hypothetical goal | "To turn on the light, push the red button." |
| 13. | Learning an operator to perform in a hypothetical state: general policy (active at all times) | "If the light is bright, then dim the light." |
| 14. | Learning an operator to perform in a hypothetical state: contingency within a particular procedure | "If the block is metal, then grasp the magnet" to `pick up` |
| 15. | Learning operator effects | pushing the red button turns on the light |
| 16. | Learning non-perceivable operator effects and associated inferences to recognize them | the magnet becomes `stuck-to` a metal block when moved above it |
| 17. | Learning control knowledge: learning which of a set of operators to prefer | two ways to grasp a small metal block |
| 18. | Learning control knowledge: learning operators are indifferent | two ways to grasp a small metal block |

Table 6: Instructional capabilities demonstrated by INSTRUCTO-SOAR.

## 8.1 Empirical Evaluation

Most empirical evaluations of machine learning systems take one of four forms, each appropriate for addressing different evaluation questions:

A.  *Comparison to other systems.* This technique is useful for evaluating how overall performance compares to the state of the art. It can be used when there are other systems available that do the same learning task.

B.  *Comparison to an altered version of the same system.* This technique evaluates the impact of some component of the system on its overall performance. Typically, the system is compared to a version of itself without the key component (sometimes called a "lesion study").

C.  *Measuring performance on a systematically generated series of problems.* This technique evaluates how the method is affected by different dimensions of the input (e.g., noise in training data).

D.  *Measuring performance on known hard problems.* Known hard problems provide an evaluation of overall performance under extreme conditions. For instance, concept learners' performance is often measured on standard, difficult datasets.

These evaluation techniques have been applied in limited ways to INSTRUCTO-SOAR. They are difficult to apply in great depth for two reasons. First, whereas most machine learning efforts concentrate on depth of a single type of learning from a single type of input, tutorial instruction requires a breadth of learning from a range of instructional interactions. Whereas depth can be measured by quantitative performance, breadth is measured by (possibly qualitative) coverage – here, our coverage of 7 out of 11 instructability requirements. Second, tutorial instruction has not been extensively studied in machine learning, so there is not a battery of standard systems and problems available. Nonetheless, evaluation techniques (B), (C), and (D) have been applied to INSTRUCTO-SOAR to address specific evaluation questions:

B.  *Comparison to altered version:* We removed frame-axiom knowledge to illustrate the effect of prior knowledge on the agent's performance, as described in Section 6.5. Without prior knowledge, the agent is unable to explain instructions and must resort to inductive methods. Thus, removing frame-axiom knowledge increased the amount of instruction required and reduced learning quality. We also compared versions of the agent that use different instruction recall strategies (Section 6.3).

C.  *Performance on systematically varied input:* We examined the effects of varying three dimensions of the instructions given to the agent. First, we compared learning curves for instruction sequences of different lengths (Section 6.2). As the graphs in Figure 10 show, INSTRUCTO-SOAR's execution time for an instructed procedure varies with the number of instructions in sequence used to teach it. Total execution time drops each time the procedure is executed, according to a power law function, until the procedure has been learned in general form. Second, we compared teaching a procedure through hierarchical subtasks versus using a flat instruction sequence. Based on the power law result, we predicted that hierarchical instruction would allow faster general learning than flat instruction. This prediction was confirmed empirically. Third, we examined the number of instruction orderings that can be used to teach a given procedure to Instructo-Soar in order to measure the value of supporting command flexibility. Rather than an experimental measurement, we performed a mathematical analysis. The analysis showed that due to command flexibility, the number of instruction sequences that can be used to teach a given procedure is very large, growing exponentially with the number of primitive steps in the procedure (Huffman, 1994).

D. *Performance on a known hard problem:* Since learning from tutorial instruction has not been extensively studied in machine learning, there are no standard, difficult problems. We created a comprehensive instruction scenario by crossing the command flexibility, situation flexibility, and knowledge-type flexibility requirements. The scenario, described in detail in (Huffman, 1994), contains 100 instructions and demonstrates 17 of Instructo-Soar's 18 instructional capabilities from Table 6 (it does not include learning indifference in selecting between two operators). The agent learns about 4,700 chunks during the scenario, including examples of each type of PSCM knowledge, that extend the agent's domain knowledge significantly.

## 9. Limitations and Further Research

This work's limitations fall into three major categories: limitations to tutorial instruction as a teaching technique, limitations of the agent's general capabilities, and limitations because of incomplete solutions to the mapping, interaction, and transfer problems. We discuss each of these in turn.

### 9.1 Limitations of Tutorial Instruction

Tutorial instruction is both highly interactive and situated. However, much of human instruction is either non-interactive or unsituated (or both), and these have not been considered in this work. In non-interactive instruction, the content and flow of information to the student is controlled primarily by the information source. Examples include classroom lectures, instruction manuals, and textbooks. One issue in using this type of instruction is locating and extracting the information that is needed for particular problems (Carpenter & Alterman, 1994). Non-interactive instruction can contain both situated information (e.g., worked-out example problems, Chi et al., 1989; VanLehn, 1987) and unsituated information (e.g., general expository text).

Unsituated instruction conveys general or abstract knowledge that can be applied in a large number of different situations. Such general-purpose knowledge is often described as "declarative" (Singley & Anderson, 1989). For example, in physics class, students are taught that $F = m \cdot a$; this general equation applies in specific ways to a great variety of situations. The advantage of unsituated instruction is precisely this ability to compactly communicate abstract knowledge that is broadly applicable (Sandberg & Wielinga, 1991).

However, to use such abstract knowledge, students must learn how it applies to specific situations (Singley & Anderson, 1989).

## 9.2 Limitations of the Agent

An agent's inherent limitations constrain what it can be taught. We have developed our theory of learning from tutorial instruction within a particular computational model of agents (the PSCM), and within this computational model, we implemented an agent with a particular set of capabilities to demonstrate the theory. Thus, both the weaknesses of the computational model and the specific implemented agent must be examined.

### 9.2.1 Computational Model

The problem space computational model is well suited for situated instruction because of its elements' close correspondence to the knowledge level (facilitating mapping from instructions to those elements), and its inherently local control structure. However, the PSCM's local application of knowledge makes it difficult to learn global control regimes through instruction, because they must be translated into a series of local decisions that will each result in local learning.

A second weakness of the PSCM is that it provides a theory of the functional *types* of knowledge used by an intelligent agent, but gives no indication of the possible *content* of that knowledge. A content theory of knowledge would allow a finer grained analysis of an agent's instructability, within the larger-grained knowledge types analysis provided by the PSCM.

### 9.2.2 Implemented Agent's Capabilities

Producing a definitive agent has not been the goal of this work. Rather, the Instructo-Soar agent's capabilities have been developed only as needed to demonstrate its instructional learning capabilities. Thus, it is limited in a number of ways.[16] For instance, it performs simple actions serially in a static world. This would not be sufficient for a dynamic domain such as flying an airplane, where multiple goals at multiple levels of granularity, involving both achievement and/or maintenance of conditions in the environment, may be active at once (Pearson et al., 1993). Instructo-Soar's procedures are implemented by a series of locally decided steps, precluding instruction containing procedure-wide (i.e., non-local) path constraints (e.g., "Go to the other room, but don't walk on the carpeting!"). There is only a single agent in the world, precluding instructions that involve cooperation with other agents (e.g., two robots carrying a couch) and instructions that require reasoning about other agents' potential actions (e.g., "Don't go down the alley, because your enemy may block you in.")

The agent has complete perception (clearly unrealistic in real physical domains), so it never has to be told where to look, or asked to notice a feature that it overlooked. In contrast, our instruction protocols show that human students are often told where to attend or what features to notice. Instructo-Soar's world is noise-free, so the agent does not need to reason or receive instruction about failed actions. Because it has complete perception and a noise-free environment, the agent does not explicitly reason about uncertainty in its perceptions or actions, and we have not demonstrated handling instructions that explicitly describe uncertain or probabalistic outcomes.[17] The agent also does not reason about time (as, e.g., Vere and Bickmore's (1990) Homer does), so it cannot be taught to perform tasks in a time-dependent way. It does not keep track of states it has seen or actions it performs (other than its episodic instruction memory), so it cannot be asked to "do what you did before." Similarly, it cannot learn procedures that are defined by a particular sequence of actions, rather than a set of state conditions to achieve. For example, it cannot be taught how to dance, because dancing does not result in a net change to the external world. Finally, whenever the agent does not know what to do next, it asks for more instruction. It never tries to determine a solution through search and weak methods such as means-ends analysis. Adding this capability would decrease its need for instruction.

---

16. These limitations are of the particular agent implemented here, not of Soar, which has been used to build more powerful agents (e.g., Jones et al., 1993; Pearson et al., 1993).

In addition to the agent's capabilities, INSTRUCTO-SOAR is limited because its solutions to the mapping, interaction, and transfer problems are incomplete in various ways. These limitations are discussed next.

## 9.3 Mapping Problem

INSTRUCTO-SOAR employs a straightforward approach to mapping instructions into the agent's internal language, and leaves all of the problems of mapping difficult natural language constructions unaddressed. Some of the relevant problems include reference resolution, incompleteness, and the use of domain knowledge in comprehension. Mapping can even require further instruction, as in this interaction to resolve a referent:

> > Grab the explosive block.
> > *Which one is that?*
> > > The red one.

This type of interaction is not supported by INSTRUCTO-SOAR.

In addition to these general linguistic problems, INSTRUCTO-SOAR makes only limited use of semantic information when learning new operators. For example, when it first reads "Move the red block left of the yellow block," it creates a new operator, but does not make use of the semantic information communicated by "Move...to the left of." A more complete agent would try to glean any information it could from the semantics of an unfamiliar command.

## 9.4 Interaction Problem

The agent's shortcomings on the interaction problem center around its three requirements: ($I_1$) flexible initiation of instruction, ($I_2$) full flexibility of knowledge content, and ($I_3$) situation flexibility. ($I_1$): In INSTRUCTO-SOAR, instruction is initiated only by the agent.

---

17. In the instruction protocols we analyzed, most instructions were *incomplete* (missing conditions like those INSTRUCTO-SOAR learns), but rarely described uncertainty explicitly.

This limits the instructor's ability to drive the interaction or to interrupt the agent's actions with instruction: "No! Don't push that button!"[18]

($I_2$): INSTRUCTO-SOAR provides flexibility for commands, but not for instructions that communicate other kinds of information. Similar to the notion of discourse coherence (Mann & Thompson, 1988), a fully flexible tutorable agent needs to support any instruction event with *knowledge* coherence; that is, any instruction event delivering knowledge that makes sense in the current context. The great variety of knowledge that could be relevant at any point makes this requirement difficult.

($I_3$): INSTRUCTO-SOAR provides situation flexibility by handling both implicitly and explicitly situated instructions, but hypothetical situations can only be referred to within a single instruction. Human tutors often refer to one hypothetical situation over the course of multiple instructions.

## 9.5 Transfer Problem

This work has focused primarily on the transfer problem – producing general learning from tutorial instruction – and most of its requirements have been met. However, the inductive heuristics that INSTRUCTO-SOAR uses are not very powerful.

In addition, two transfer problem requirements have not been achieved. First, ($T_7$) INSTRUCTO-SOAR has not yet demonstrated instructional learning in coexistence with learning from other knowledge sources. Nothing in INSTRUCTO-SOAR's theory precludes this coexistence, however. Learning from other knowledge sources could be invoked and possibly enhanced through instruction. For instance, an instructor might invoke learning from observation by pointing to a set of objects and saying "This is a tower"; similarly, an instruction containing a metaphor could invoke analogical learning. One application where instruction could potentially enhance other learning mechanisms is within "personal assistant" software agents that learn by observing their users (e.g., Maes, 1994; Mitchell et al., 1994). Adding the ability to learn from verbal instructions in addition to observations would allow users to explicitly train these agents in situations where learning from observation alone may be difficult or slow.

Second, ($T_6$) INSTRUCTO-SOAR cannot recover from incorrect knowledge that leads to either invalid explanations or incorrect external performance. Such incorrect knowledge may be a part of the agent's initial domain theory, or may be learned through faulty instruction. Inability to recover from incorrect knowledge precludes instruction by general case and exceptions; for instance, "Never grasp red blocks," and then later, "It's ok to grasp the ones with safety signs on them." In order to avoid learning anything incorrect, whenever INSTRUCTO-SOAR attempts to induce new knowledge, it asks for the instructor's verification before adding the knowledge to its long-term memory. Human students do not ask for so much verification; they appear to jump to conclusions, and alter them later if they prove to be incorrect based on further information.

Rather than always verifying knowledge being learned, our next generation of instructable agents will learn from reasonable inferences without verification (although they may ask for verifications in extreme cases). We have recently produced such an agent (Pearson &

---

18. We have recently added a simple interruptability capability to a new version of INSTRUCTO-SOAR that incorporates recovery from incorrect knowledge (Pearson & Huffman, 1995).

Huffman, 1995) that incorporates current research on incremental recovery from incorrect knowledge (Pearson & Laird, 1995). This agent learns to correct overgeneral knowledge that it infers when completing explanations of instructions. The correction process is triggered when using the overgeneral knowledge results in incorrect performance (e.g., an action that the agent expects to succeed does not). In the long run, we believe this work could push research on incremental theory revision and error recovery, because instructable agents can be taught many types of knowledge that may need revision.

## 10. Conclusion

Although much work in machine learning aims for *depth* at a particular kind of learning, INSTRUCTO-SOAR demonstrates *breadth* – of interaction with an instructor to learn a variety of types of knowledge – but all arising from one underlying technique. This kind of breadth is crucial in building an instructable agent because of the great variety of instructions and the variety of knowledge that they can communicate. Because instructable agents begin with some basic knowledge of their domain, INSTRUCTO-SOAR uses an analytic, explanation-based approach to learn from instructions, which makes use of that knowledge. Because instructions may be either implicitly or explicitly situated, INSTRUCTO-SOAR situates its explanations of each instruction within the situation indicated by the instruction. Finally, because the agent's knowledge is often deficient for explaining instructions, INSTRUCTO-SOAR employs four different options for dealing with incomplete explanations, and selects between these options dynamically depending on the instructional context.

Because of its availability and effectiveness, tutorial instruction is potentially a powerful knowledge source for intelligent agents. INSTRUCTO-SOAR illustrates this in a simple domain. Realizing instruction's potential in fielded applications will require more linguistically able agents that incorporate robust techniques for not only acquiring knowledge from instruction, but also refining that knowledge as needed based on performance and further instruction.

## Acknowledgements

This work was performed while the first author was a graduate student at the University of Michigan. It was sponsored by NASA/ONR under contract NCC 2-517, and by a University of Michigan Predoctoral Fellowship. Thanks to Paul Rosenbloom, Randy Jones, and our anonymous reviewers for helpful comments on earlier drafts.